\documentclass[10pt, a4paper]{article}

\usepackage[]{lrec-coling2024} % this is the new style
\usepackage{dcolumn}
\usepackage{booktabs}
\usepackage{multirow}
\usepackage{graphicx}
\usepackage{amsmath}
\usepackage{enumitem}
\usepackage{array}
\usepackage{xcolor}
\usepackage{algorithm,algorithmic}
\usepackage{caption}
\usepackage{subcaption}
\usepackage{wrapfig}
\usepackage[nohyperlinks,nolist]{acronym}

\usepackage[nohyperlinks,nolist]{acronym}
\newcommand{\tx}[1]{``\textit{#1}''}
\newcolumntype{.}{D{.}{.}{-1}}
\definecolor{mygreen}{RGB}{21,71,52}
\newcommand{\fixes}[2]{\textcolor{black}{#2}}
\title{\textbf{Unveiling Vulnerability of Self-Attention}}
\name{Khai Jiet Liong\textsuperscript{1,2\sthanks{\;\;Equal Contribution}},
Hongqiu Wu\textsuperscript{1,2}, 
Hai Zhao\textsuperscript{1,2\sthanks{\;\;Corresponding author; This paper was partially supported by Joint Research Project of Yangtze River Delta Science and Technology Innovation Community (No. 2022CSJGG1400).}}, 
Li Xiaoshan\textsuperscript{4*}, 
Bian Minjie\textsuperscript{3,4}}

\address{$^1$ Department of Computer Science and Engineering, Shanghai Jiao Tong University \\
        $^2$ Key Laboratory of Shanghai Education Commission for Intelligent Interaction\\
        and Cognitive Engineering, Shanghai Jiao Tong University\\
        $^3$ Shanghai Technology Innovation Department, Shanghai Data Group Co., Ltd \\
        $^4$ Shanghai Data Group Co., Ltd \\
        \{liongkj, wuhongqiu\}@sjtu.edu.cn\\
        zhaohai@cs.sjtu.edu.cn\\
        lixs@sdata.net.cn\\
        bianmj@sdata.net.cn}

\abstract{
Pre-trained language models (PLMs) are shown to be vulnerable to minor word changes, which poses a big threat to real-world systems. While previous studies directly focus on manipulating word inputs, they are limited by their means of generating adversarial samples, lacking generalization to versatile real-world attack. This paper studies the basic structure of transformer-based PLMs, the self-attention (SA) mechanism.
(1) We propose a powerful perturbation technique \textit{HackAttend}, which perturbs the attention scores within the SA matrices via meticulously crafted attention masks. We show that state-of-the-art PLMs fall into heavy vulnerability that minor attention perturbations $(1\%)$ can produce a very high attack success rate $(98\%)$. Our paper expands the conventional text attack of word perturbations to more general structural perturbations. (2) We introduce \textit{S-Attend}, a novel smoothing technique that effectively makes SA robust via structural perturbations. We empirically demonstrate that this simple yet effective technique achieves robust performance on par with adversarial training when facing various text attackers.
    Code is publicly available at \url{github.com/liongkj/HackAttend}.
 \\ \newline  }

\begin{document}

\maketitleabstract

\section{Introduction}
Pre-trained language models (PLMs), e.g. BERT \citep{DBLP:conf/naacl/DevlinCLT19}, GPT \citep{radford_improving_2018}, RoBERTa \citep{DBLP:journals/corr/abs-1907-11692}, ALBERT \citep{DBLP:conf/iclr/LanCGGSS20}, and DeBERTa \citep{DBLP:conf/iclr/HeLGC21} have demonstrated human-level performances on a series of challenging natural language processing (NLP) tasks, e.g. reading comprehension \citep{DBLP:journals/tacl/SunYCYCC19} and logical reasoning \citep{ZHOU2020275,yu2020reclor,wu2023empower}.
Yet, despite their impressive capability, studies unveil that such deep neural networks can be easily misled by minor word perturbations, which poses a significant challenge in deploying robust NLP systems.

Augmenting training data with adversarial samples generated by text attack has been proven to be an effective technique to create robust language models.
Existing attackers generate the adversarial samples by manipulating the input text, e.g. by word substitution, swapping or insertion \citep{DBLP:conf/aaai/JinJZS20}.
Nevertheless, these defense methods are limited by the means of the attack and have limitations in their effectiveness against attacks in more general situations, such as word substitutions beyond synonyms or semantics.
Moreover, incorporating such adversarial samples into training always results in a large degradation of the performance.

Our key vision is that the vulnerability of a language model derives from its architecture and inner mechanism.
By examining the vulnerability of the inner mechanism, it is possible to gain insights into why language models are susceptible to input perturbations.
In this paper, we specifically investigate the self-attention (SA) mechanism in the context of model robustness, an area that has received limited attention despite its fundamental role in PLMs.

% In NLP, adversarial attacks can have severe consequences. For instance, an attacker can manipulate sentiment analysis models to provide incorrect sentiment predictions, potentially leading to misinformation or financial losses. Additionally, adversarial attacks, specifically poison attack can be used to inject backdoor triggers into models, leading to backdoor attacks \citep{garg2020can,zhang2021advdoor, Kurita2020Weight,guo2021gradient}. Furthermore, with the popularity of large language models (LLMs), studies have revealed attacks such as prompt injection to misalign and bypass safety checks \citep{perez2022ignore}. Moreover, variants of these attack strategies have been extensively studied in image and speech data \citep{madry2017towards,carlini2018audio}.

\begin{figure}
    \centering
    \includegraphics[width=0.9\linewidth]{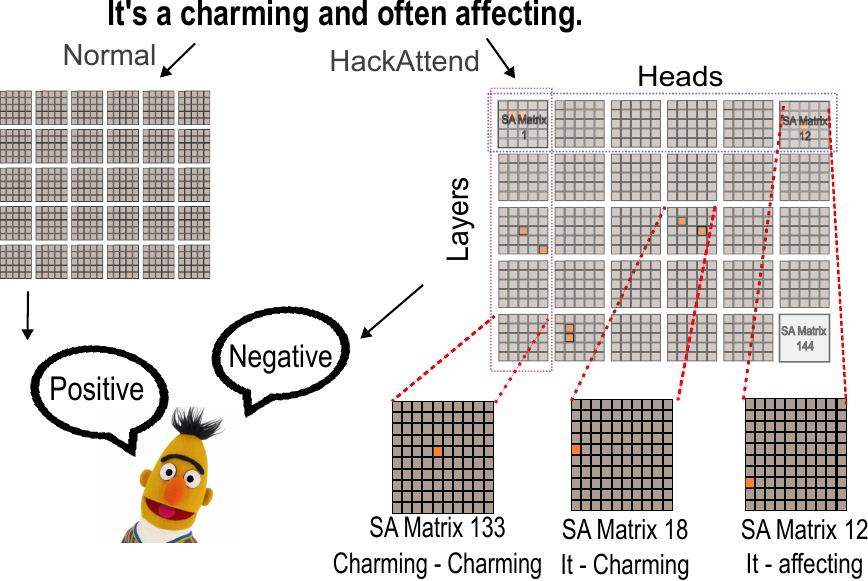}
    \caption{
    % \textit{HackAttend} perturbs (flips from 1 to 0) the SA unit (colored in orange) of SA matrices to induce misclassification. The sample is taken from \textsc{SST-2}, a sentimental analysis dataset.
    \fixes{rebuttal}{Illustration of the process of \textit{HackAttend} manipulating the SA mechanism. The perturbation of SA units (highlighted in orange) demonstrates how the algorithm induces misclassification in sentiment analysis by flipping the activation states. The example perturbation shows the transition from a positive to a negative sentiment interpretation, providing a visual representation of \textit{HackAttend}'s effect on the model's decision-making process for the SST-2 dataset.}
    }
    \label{fig:hackattend}
\end{figure}

% Most existing white-box attack methods in NLP focus primarily on the input embedding layer, neglecting potential vulnerabilities in other model components. In particular, the self-attention mechanism (SA) in transformer models has received limited attention in the context of adversarial attacks. SA blocks, which are fundamental components of the multi-layer architecture in transformer models, enable the model to attend to different parts of the input sequence simultaneously, capturing dependencies and learning representations. Recent research has revealed the utilization of adversarial training to exploit specific features and spurious tokens within self-attention mechanisms, thereby bolstering the model's robustness \citep{wu2022adversarial}. By comprehending and leveraging the vulnerabilities inherent in the SA mechanism, new insights into adversarial attacks in NLP can be gained.

To examine the vulnerability of SA in PLMs, We first propose a novel perturbation strategy \textit{HackAttend}. Unlike previous attack methods which focus on the \textbf{input words}, our algorithm perturbs the SA weights within the \textbf{SA matrices} to trigger the model to a wrong prediction.

Figure~\ref{fig:hackattend} illustrates how \textit{HackAttend} generates adversarial samples in the form of custom attention mask which successfully disrupt SA mechanism's ability to capture contextual information effectively. Empirical experiments were conducted on a wide range of tasks, including reading comprehension, logical reasoning, and sentiment analysis, and natural language inference. Our results demonstrate that state-of-the-art language models are heavily vulnerable to \textit{HackAttend}, achieving high attack success rate with only minor SA perturbations.

Building upon our findings, we then propose a novel smoothing technique \textit{S-Attend}, which smooths the attention scores during training. Since our technique perturbs the model structure, resulting in no bias to input distribution, it unlocks impressive robustness outcome with minor performance compromise.

Our contributions are summarized below:

$\bullet$ We are the first to discuss the perturbations and smoothing technique orienting SA.

$\bullet$ We analyze the impact of different perturbation levels in the SA matrices on downstream tasks, providing insights into the role of SA in capturing task-specific signals.

$\bullet$ We propose a new smoothing technique to defend against general attacks.

% $\bullet$ We introduce Hamming Distance as metric to measure the similarity of self-attention block, acting as an indicator of perturbation strength.

% \input{sections/related.tex}
\section{Related Work}

Our work introduces a novel perturbation strategy on the SA component. Similar to the end goal of text adversarial attack algorithms, the objective of \textit{HackAttend} perturbation is to induce misclassifications in language models. Examples of text adversarial attacks includes character level attacks (e.g. TextBugger \citep{gao2018black}, DeepWordBug \citep{li2018textbugger}, HotFlip \citep{ebrahimi2018hotflip}) and the word level (e.g. TextFooler \citep{DBLP:conf/aaai/JinJZS20}, BERT-Attack \citep{li2020bertattack}, SemAttack \citep{wang2022semattack}, PWWS \citep{ren2019generating}, and BBA \citep{DBLP:conf/icml/LeeMLS22}.
While these methods pay attention on the scope of input text and word embedding, our research, in contrast, places emphasis on the architecture of the model itself. Our goal is not to propose a new algorithm for real-world adversarial attacks, but rather to gain a deeper understanding of this unexplored area.

Our work focuses on the fundamental architecture, self-attention (SA) mechanism \citep{DBLP:conf/nips/VaswaniSPUJGKP17} of PLMs such as \texttt{RoBERTa} \citep{DBLP:journals/corr/abs-1907-11692}, \texttt{ALBERT} \citep{DBLP:conf/iclr/LanCGGSS20}, and \texttt{DeBERTa} \citep{DBLP:conf/iclr/HeLGC21}.
SA has been extensively studied in the literature \citep{shi2021sparsebert,you2020hard,zhang2020sg}.
\citet{wu2022adversarial} initially discuss the robustness of SA, highlighting the model's tendency to overemphasize certain spurious keywords while ignoring overall semantics. Drawing upon this observation, we leverage this insight to develop our perturbation strategy.

In previous literature, various metrics have been employed to impose constraints on perturbation levels in order to minimize perceptibility between original and adversarial samples. Metrics such as Jaccard similarity, cosine similarity, Earth Mover's Distance \citep{Rubner2000earthmover}, MoverScore \citep{zhao2019moverscore} and BERTScore \citep{zhang2019bertscore} have proven effective in evaluating semantic similarity of the input. However, the Hamming Distance provides an alternative perspective in our evaluation, allowing us to control perturbation levels at a structural level.

It is worth noting that adversarial training is usually adopted as the means to enhance the robustness of models \citep{goodfellow2014explaining,DBLP:conf/iclr/ZhuCGSGL20,wu2023toward}. In contrast to prior research, we distinguish ourselves by removing the adversarial nature and presenting an efficient smoothing technique in improving model robustness in our study. \fixes{rebuttal}{Our method employs random masking of attention units, contrasting with conventional techniques that mask whole unimportant attention heads during training \citep{DBLP:conf/emnlp/BudhrajaPNKK20} and inference \citep{DBLP:conf/naacl/CaoW21}, or adding an extra layer of complexity \citep{DBLP:conf/naacl/FanGLWWJDZH21}.} Through empirical evidence, we demonstrate the substantial advantages of incorporating structural perturbations to achieve comprehensive and resilient robustness against diverse attacks. 

\section{Introduction to HackAttend}
\label{sec:methodology}
This section elaborates on our proposed \textit{HackAttend} perturbation design. \textit{HackAttend} distinguishes itself from conventional text attacks as our main purpose is for interpretability of the robustness of language models.

\subsection{Notations}
\begin{acronym}[TDMA]
    \small
    \acro{mask-percent}[$\alpha$]{masking percentage}
    \acro{num_layers}[$N_L$]{number of layers in the victim model}
    \acro{num_heads}[$N_H$]{number of heads in the victim model}
    \acro{nl}[$l_{max}$]{maximum number of layers perturbed}
    \acro{nh}[$h_{max}$]{maximum number of heads perturbed}
    \acro{sa_block}[SA]{self-attention block with matrix of size $N \times N$, where $N$ represents the number of non-padding tokens in the input sequence}
    \acro{sa_i_j_k}[$SA_{i,j,k}$]{the $k$\textsuperscript{th} unit in the SA block of the $j$\textsuperscript{th} head in the $i$\textsuperscript{th} layer of the model}
    \acro{M}[$M$]{original attention mask }
    \acro{M'}[$M'$]{perturbed attention mask}
    % \acro{gradientbase}[GAIR]{Gradient-based Attention Importance Ranking}
    \acro{score}[$S(\cdot)$]{importance score of a layer or head}

\end{acronym}
We first define the notations used in the methodology:

\begin{description}[style=multiline,leftmargin=2cm,font=\textbf\itshape\space,itemsep=1pt,]
    \item[$\alpha$]{Masking percentage}
    \item[$N_L$ / $N_H$]{Number of layers / heads in the victim model}
    \item[$l_{\text{max}}$ / $h_{\text{max}}$]{Maximum number of layers / heads perturbed}
    \item[$SA_{i,j,k,l}$]{The $k$\textsuperscript{th}-row $l$\textsuperscript{th}-column attention unit in the SA matrix of the $j$\textsuperscript{th} head and $i$\textsuperscript{th} layer}
    \item[$M$ / $M'$]{Original / adversarial attention mask}
    \item[$S_L$ / $S_H$]{Importance score of a layer or head}
\end{description}

\subsection{Overview of Hackattend}
\label{sub:attack_algo}
% To exploit the vulnerability of the victim model effectively, our attack algorithm prioritizes SA blocks based on their impact. A SA block refers to the self attention matrix present in a particular head and layer of the model. This SA block is represented by a matrix of size $N \times N$, where $N$ represents the number of non-padding tokens in the input sequence. Each element within the matrix represents the attention score between two tokens, ranging from 0 (indicating no attention) to 1 (indicating full attention). We adopt a greedy approach, gradually increasing the attack strength while targeting the most sensitive components of the model.

% The importance scores obtained during the process of masking entire layers and individual heads within the sensitive layers play a crucial role in the subsequent step known as the \ac{gradientbase} (details in Section ~\ref{sub:select_sa}), where critical SA units are identified and masked based on their importance scores. This step generates a perturbed mask configuration specifically designed for the attack, allowing us to explore the search space effectively while prioritizing the most influential blocks.
A PLM consists of multiple SA layers, each with varying numbers of attention heads (also referred as SA matrices). Each cell within these matrices represents a SA unit. For example, a \verb|BERT-base| model typically includes 12 (layers) x 12 (heads) attention matrices.
% Our proposed method, called \textit{HackAttend} could search for adversarial samples for any model configuration efficiently, based on the idea of greedy.
Our proposed method, \textit{HackAttend} efficiently identifies adversarial samples across various model configurations by employing a greedy algorithm approach.

The overall objective of the \textit{HackAttend} algorithm is to rank and assess the significance of every SA matrices in the model based on their eventual impact the final model prediction (discussed in Section~\ref{sub:select_layer}. We employed a novel gradient-based technique (details in Section~\ref{sub:select_sa}) to prioritize the most significant SA units within the candidate SA matrices.
We will then apply a masking operation on the prioritized SA units within a constraint requirement defined in the Section~\ref{sub:select_contraint} to ensure minimal perturbation on the attention mask while effectively exploring the search space.

Unlike traditional adversarial samples, \textit{HackAttend} generates an adversarial structure (in the form of custom attention mask) which perturbs the underlying attention calculation.

\subsection{Layers and Head Selection}
\label{sub:select_layer}
\paragraph{Layers Selection} 
% To determine the importance score of each layer, we employ an iterative process of masking all heads in the $i$\textsuperscript{th} layer $L_i \in (L_0, \ldots, L_{N_l-1})$.
% % Subsequently, we assess the impact on logits when masking $i$\textsuperscript{th} layer.

% If masking the $i$\textsuperscript{th} layer results in an incorrect prediction, we assign the highest importance score to that layer. In cases of consistent prediction, we construct a ranking based on $p(x\mid\dots)$. The formulation of logits when the SA matrices in the $i$\textsuperscript{th} layer are masked as follows:
% \begin{equation}
%     S_L[i] \propto 1-p(x\mid \rm Layer_i \; is \; masked)
% \end{equation}
% where $p(x\mid \rm Layer_i \; is \; masked)$ represents the output probability distribution of the model when the SA matrices in the $i$\textsuperscript{th} layer $SA_{i,0}, SA_{i,1}$. $\dots$, $SA_{i,{N_H}}$ are masked.
% The drop in logits serves as an indicator of the layer's importance, with a larger drop suggesting greater significance.

In our approach, we assess each layer's importance in the model by iteratively masking all attention heads within that layer. If masking a specific layer, denoted as the $i$-layer, results in an incorrect model prediction, this layer is deemed highly important and assigned a top importance score. For cases where the model still predicts correctly, layer importance is determined by the reduction in the model's output probability. A greater decrease in this probability signifies a higher importance during prediction.

% \subsection{Heads Selection}
% \label{sub:select_head}
\paragraph{Heads Selection} Similar to the process for layer selection, we use a comparable strategy to evaluate the importance of individual heads, denoted as $H_j$, where each $H_j$ corresponds to a specific head in the model's $i$\textsuperscript{th} layer. We systematically mask the heads in the $i$\textsuperscript{th} layer, one at a time, and calculate the model's output probability, resulting in the head importance score $S_H[i,j]$.
The scores were ranked in descending order, prioritizing the more vulnerable head as the target for the perturbation.
% The set of heads is denoted as $[H_0, H_1, \dots, H_{N_h-1}]$.
% which is formulated as follows:
% \begin{equation}
%   S_H[i,j] \propto 1 - p(x\mid \rm Head_j \; in \; Layer_i \; is \; masked)
% \end{equation}

% rank the heads in descending order based on their importance scores.
% The confidence score $S(SA_{i,j})$ represents the probability of the ground truth label when the attention weights of head $H_j$ in layer $L_i$ are masked. $\mathbf{p}(x \mid M'_{i,j})$ denotes the output probability distribution of the model when $SA_{i,j}$ matrix is masked.

\subsection{SA Units Selection}
\label{sub:select_sa}
% \begin{wrapfigure}{o}{0.15\textwidth}
%     \includegraphics[width=\linewidth]{figures/hook.pdf}
%     \caption{Placement of \\mask hook}
% \end{wrapfigure}

To determine the SA units that disrupt the model's attention mechanism, we utilize a gradient-based algorithm for ranking the SA units, inspired by the work of \citet{wu2022adversarial} where the underlying premise is SA units with larger gradients has more significant impact on model predictions.
By masking these SA units in the worse-case direction, we maximizes the empirical training risk.
\begin{figure}[h]
    \centering
    \includegraphics[width=0.38\linewidth]{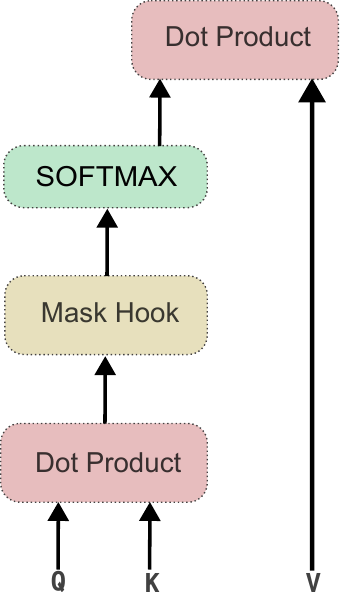}
    \caption{Placement of mask hook}
    \label{fig:hook}
\end{figure}
Our process involves the use of \ac{M}, where no attention units are initially masked at the start of each forward step.
We then compute the gradients of the loss function with respect to the \ac{M}, and rank the SA units based on their gradients in the backward step.

To obtain the gradients of SA units during the backward pass, we add a custom backward hook in the encoder layer, illustrated in Figure~\ref{fig:hook}, to capture the gradients flowing through \ac{M}.
These modifications are applied to the original SA layer, specifically just before the softmax operation, while keeping the remaining components unchanged.

% one more forward to get the gradients
% mask

Nevertheless, it has come to our attention that specific attention heads may generate very small gradients in certain samples such that ranking with gradients is impractical, as these gradients may essentially become zero across all units. As a fallback strategy, the attention scores obtained during the forward pass will serve as a proxy score as proposed by \citet{michel2019sixteen}.
Furthermore, an ablation study comparing both ranking strategies can be found in Subsection~\ref{sub:effect_of_importance_ranking_strategy_for_sa_unit}.

% \subsection{Setup}

\subsection{Constraints}
\label{sub:select_contraint}
In the final step of the algorithm, the attention mask \acs{M} with the highest importance rankings within the candidate SA matrices is perturbed, preventing the model from attending to those units during computations.
% (\acs{M})
In this process, masking involves expanding the original attention mask, which has
dimensions of \texttt{[seq\_len, seq\_len]}, to create a modified mask with dimensions of \texttt{[num\_layers, num\_head, seq\_len, seq\_len]}, and selectively deactivating specific SA units.

To ensure the learning process remains intact, a constraint is imposed on the percentage of tokens to be perturbed. This constraint guarantees the high similarity between $M$ and $M'$, thereby preserving the overall model's functionality while maintaining an optimal adversarial approach.

To implement the token masking constraint, the parameter \acl{mask-percent}, denoted as \acs{mask-percent}, is introduced. It is assigned a value of $0.01$, representing $1.0\%$ of the non-padding tokens. For example, let's consider a SA matrix with 225 units (a $15 \times 15$ matrix), and the goal is to mask $1\%$ of the units. In this case, the top 2 units with the highest gradients would be selected and denoted as ${SA_{i,j,k_1,l_1}}$ and ${SA_{i,j,k_2,l_2}}$, where $(k_1,l_1)$ and $(k_2,l_2)$ represent the indices of the top 2 sorted SA units.
$M'$ serves as a replacement for the original attention mask before it was fed into the victim model.
% $M'$ is the generated adversarial structure, which is then evaluated.

\subsection{Algorithm}
\textit{HackAttend} employs a greedy approach to generate adversarial attention mask by ranking all SA matrices. It iteratively selects the most vulnerable layers and heads based on their importance scores and searches for important SA units based on gradients. The resulting adversarial structure, denoted as $M'$, is then assessed whether it could induce misclassification of the victim model.
The algorithm, with its greedy decision-making characteristic, is presented in Algorithm~\ref{algo:greedy_attack}.

\begin{algorithm}[h]
    \small

    \begin{algorithmic}[1]

        % maxumum number of layers attack
        \REQUIRE{\Ac{nl}, \Ac{nh}, Victim model ($f$), \Ac{mask-percent}}
        % \ENSURE{Attack Status}
        \STATE Select a sample $\{x,y\}$
        % \FOR{Sample $x_i$ in $x$}
        % \STATE $S_L \gets \textsc{RankLayers}(x_i)$
        \STATE Go forward step and obtain the gradients
        \STATE $S_L \gets $ Rank all layers based on importance scores
        \FOR{$l_i \gets 1$ \TO \ac{nl}}
        % \STATE \textsc{LayersCan} $\gets  S_L[:l_i]$ \COMMENT{Select the top $l_i$ sensitive layers}
        \STATE $S_H \gets $ Rank the heads in the $S_L[l_i]$\textsuperscript{th} layer based on their importance scores
        \FOR{$h_j \gets 1$ \TO \ac{nh}}
        % \STATE $\textsc{HeadsCan} \gets  S_H[:h_i]$ \COMMENT{Select the top $h_i$ important heads}
        \STATE $M' \gets $ Mask the units in $S_H[h_j]$\textsuperscript{th} head with top $\alpha$ largest gradients
        \STATE $\hat{y} \gets f(x, M')$
        \IF{$\hat{y} \ne y$}
        \RETURN SUCCESS
        \ENDIF
        \ENDFOR
        \ENDFOR
        \RETURN FAIL
        % \ENDFOR

        \caption{\textit{HackAttend}}
        \label{algo:greedy_attack}
    \end{algorithmic}
\end{algorithm}

\section{Empirical Experiment}
\label{sec:results}
This section reports our empirical results.
\subsection{Evaluation Settings}
\label{sec:eval_settings}
We evaluate the effectiveness of the perturbations and its impact using the following metrics:

\paragraph{Hamming Distance} This metric quantifies the extent of perturbations to the SA matrix. Since the attention matrix is a binary matrix, Hamming distance measures the number of bits that differ between $M$ and $M'$. We average the Hamming distance across each perturbed SA matrix as follows:

\[
    d_H(M,M') = \frac{\sum_{i=1}^{N_{SA}} (M_i \oplus M'_i)}{N_{SA}}
\]
where $\oplus$ represents the XOR (exclusive or) operation, $N_{SA}$ is the number of SA matrices perturbed.

\paragraph{Clean Accuracy} This metric is used to measure the accuracy score on the clean set.
% \[
%   Acc_{Clean} = \frac{\text{\# correct predictions}}{\text{\# predictions}}
% \]

\paragraph{Robust Accuracy} This metric is used to measure the accuracy score under attack/perturbation.
% \[
%   Acc_{Robust} = \frac{\text{\# correct predictions after attack}}{\text{\# predictions}}
% \]

\paragraph{Attack Success Rate (ASR)} This metric is used to measure the perturbation's success rate when the model makes an incorrect prediction after the perturbation. The ASR is computed as follows:
\[
    ASR = \frac{\text{\# successful perturbation}}{\text{\# correct predictions}}
\]

An effective perturbation algorithm has the following properties:
$\,$1) a low Hamming distance, indicating high degree of similarity and the perturbation is cunning yet minimally affects the attention structure; $\,$2) a high ASR and/or low robust accuracy is desired, indicating the effectiveness successfully inducing misclassifications; 
% $\,$3) a high clean accuracy, indicating the original performance is marginally compromise;

% \input{sections/experiments.tex}
\subsection{Dataset}

For our experiments, we choose four representative NLP tasks:

$\bullet$ \textbf{Sentiment Analysis}: Stanford Sentiment Treebank (\textsc{SST-2}) \citep{DBLP:conf/emnlp/SocherPWCMNP13};

$\bullet$ \textbf{Natural Language Inference (NLI)}: \textsc{HellaSWAG} \citep{DBLP:conf/acl/ZellersHBFC19}, a multiple-choice common-sense reasoning dataset;

$\bullet$ \textbf{Dialogue Comprehension}: Dialogue-based machine reading comprehension (\textsc{DREAM}) \citep{DBLP:journals/tacl/SunYCYCC19}, in a multiple-choice format;

$\bullet$ \textbf{Logical Reasoning}: ReClor \citep{yu2020reclor}, a machine reading comprehension in a multiple-choice format.

% table to show average sequence length of test set, round to 2 dp
% dream 75.26625226279424
% alphanli 33.9182739457339
% hellaswag 86.66555467038438
% \textsc{MNLI} 39.05369332654101
% \textsc{sst}2- }5.163990825688074
\begin{table}[ht]
    \small
    \centering
    \begin{tabular}{l D{.}{.}{2.3} D{.}{.}{3.1} D{.}{.}{2.1}}
        \toprule
        \textbf{Dataset}   & \multicolumn{1}{c}{\textbf{Mean}} & \multicolumn{1}{c}{\textbf{Max}} & \multicolumn{1}{c}{\textbf{Min}} \\
        \midrule
        \textsc{DREAM}     & 75.3                              & 128.0                            & 24.0                             \\
        \textsc{HellaSWAG} & 86.7                              & 128.0                            & 19.0                             \\
        \textsc{ReClor}    & 125.1                             & 128.0                            & 66.8                             \\
        \textsc{SST-2}     & 25.2                              & 55.0                             & 4.0                              \\
        \bottomrule
    \end{tabular}
    \caption{Sequence length of the Test/Dev split.}
    \label{tab:stats}
\end{table}
\subsection{Setup}

\paragraph{Victim Models} For our experiments, we selected \texttt{BERT-base} as the victim model. We initially fine-tuned this pretrained model on a target dataset (training details can be found in Appendix), followed by perturbation experiments using the same dataset on a single NVIDIA TITAN RTX 24G.
% with hyperparameter settings shown in Appendix Table~\ref{tab:hyperparameters}.

In the \textit{HackAttend} implementation, we employ two configurations to limit the number of layers/heads that the algorithm can target. The full-scale setting uses parameters \ac{nh} and \ac{nl} set to 12, while the half-scale is set to 6, thereby reducing the search space. Note that these settings do not imply that all layers/heads are perturbed simultaneously. The attention mask is related to real sequence length of the input text i.e. the non-pad tokens. With the same mask percentage $\alpha$, longer sequences results in more SA units being masked. Hence, the statistics of sequence length is shown in Table~\ref{tab:stats}.

% This reduction balances attack effectiveness with computational efficiency. 
% We employ two configurations for. The full-scale attack has \ac{nh} and \ac{nl} set to 12, while the half-scale attack reduces the search space by setting \ac{nh} and \ac{nl} to 6, balancing attack effectiveness and computational 
\subsection{Results}

\paragraph{Main Results}

\begin{table*}[]{}
    \centering
    \small
    % \begin{tabular}{c|c|ccccclc}
    \setlength{\tabcolsep}{3pt}
    \begin{tabular}{@{}cccccccc@{}}
        \toprule
        Dataset                    & Max N & ASR$\%$ & \textit{clean}$\%$    & \textit{robust}$\%$ & \# Query & Hamming  \\
        \toprule
        \multirow{2}{*}{DREAM}     & $12$  & $98.9$  & \multirow{2}{*}{64.7} & $0.7$               & $18.6$   & $611.4$  \\
                                   & $6$   & $91.2$  &                       & $5.7$               & $11.2$   & $618.2$  \\ \midrule
        \multirow{2}{*}{HellaSWAG} & $12$  & $99.9$  & \multirow{2}{*}{39.6} & $0.0$               & $8.8$    & $1222.2$ \\
                                   & $6$   & $96.7$  &                       & $1.3$               & $7.1$    & $1232.8$ \\ \midrule
        \multirow{2}{*}{ReClor}    & $12$  & $100.0$ & \multirow{2}{*}{51.8} & $0.0$               & $7.3$    & $2151.3$ \\
                                   & $6$   & $99.6$  &                       & $0.2$               & $6.5$    & $2153.7$ \\ \midrule
        \multirow{2}{*}{SST-2}     & $12$  & $27.4$  & \multirow{2}{*}{93.9} & $67.8$              & $123.6$  & $9.3$    \\
                                   & $6$   & $10.2$  &                       & $83.8$              & $34.1$   & $9.5$    \\ \bottomrule
    \end{tabular}
    \caption{Max N indicates the maximum number of candidate heads/layers and \ac{mask-percent}= 1.0$\%$ and the results for \textsc{DREAM}, \textsc{HellaSWAG}, \textsc{ReClor}, and \textsc{SST-2} are reported on the dev set.}
    \label{t1}
\end{table*}

Table~\ref{t1} demonstrates the overall results of \textit{HackAttend} across various tasks. The introduced perturbations demonstrate its effectiveness, achieving an ASR of at least \textbf{$98.9\%$} in 3 out of 4 tasks and reducing the clean accuracy on \textsc{DREAM} ($\sim 64 \%$), \textsc{ReClor} ($\sim 52 \%$), and \textsc{HellaSWAG} ($\sim 39 \%$) under the full-scale setting.
\textit{HackAttend} is sub-optimal in the case of \textsc{SST-2} in both configurations, yielding ASR of 27.4\% and robust accuracy drop ($\sim 26 \%$) and $10.2\%$ ASR, robust accuracy drop ($\sim 10 \%$), respectively.

We hypothesize that this discrepancy can be attributed to the varying degrees of the reliance on SA layer across different tasks. For example, in simple tasks like sentiment analysis, the model depends on a combination of linguistic features and local keywords, making them less sensitive to perturbations. In contrast, tasks involving complex question answering and story comprehension heavily rely on attention mechanisms, as demonstrated empirically in Subsection~\ref{sec:vulnerability_layers}.
% to capture intricate linguistic relationships and global context. Consequently, the perturbations can effectively disrupt the model's ability to understand local details and specific phrases.

\paragraph{Results on Perturbation Strength}
Notably, in the half-scale setting, the gradient-based algorithm remains significant in most of the datasets.
On \textsc{DREAM}, the drop in ASR\% was only marginal, measuring at $7.7\%$. On \textsc{HellaSWAG}, the drop was even smaller, at $3.2\%$. The \textsc{ReClor} dataset exhibited minimal decrease in ASR, with only $0.4\%$ reduction.
These findings indicate that half-scale setting is effective, but perturbations under the full-scale setting yield more promising results.

\begin{table}[]
    \centering
    \small
    \setlength{\tabcolsep}{2pt}
    \begin{tabular}{lccc}
        \toprule
        Setting             & ASR\%              & \# Query           & Hamming               \\
        \midrule
        \textbf{DREAM}      &                    &                    &                       \\
        Baseline            & 61.7$_{\pm 1.31}$  & 76.4$_{\pm 1.25}$  & 565.3$_{\pm 79.20}$   \\
        Imp. w/o Grad       & 64.6$_{\pm 0.14}$  & 65.8$_{\pm 0.07}$  & 562.4$_{\pm 3.34}$    \\
        Grad w/o Imp.       & 98.9$_{\pm 0.00}$  & 20.0$_{\pm 0.09}$  & 611.9$_{\pm 0.33}$    \\
        \textit{HackAttend} & 98.9               & 18.6               & 611.4                 \\
        \midrule
        \textbf{HellaSwag}  &                    &                    &                       \\
        Baseline            & 88.4$_{\pm 0.06}$  & 39.9$_{\pm 0.18}$  & 1191.6$_{\pm 2.17}$   \\
        Imp. w/o Grad       & 88.8$_{\pm 0.02}$  & 33.5$_{\pm 0.01}$  & 1195.6$_{\pm 0.38}$   \\
        Grad w/o Imp.       & 99.9$_{\pm 0.00}$  & 8.3$_{\pm 0.01}$   & 1221.9$_{\pm 0.13}$   \\
        \textit{HackAttend} & 99.9               & 8.8                & 1222.2                \\
        \midrule
        \textbf{ReClor}     &                    &                    &                       \\
        Baseline            & 77.9$_{\pm 0.10}$  & 61.2$_{\pm 1.45}$  & 2123.4$_{\pm 203.71}$ \\
        Imp. w/o Grad       & 77.6$_{\pm 0.70}$  & 47.4$_{\pm 0.55}$  & 2109.3$_{\pm 21.85}$  \\
        Grad w/o Imp.       & 100.0$_{\pm 0.00}$ & 5.5$_{\pm 0.00}$   & 2151.3$_{\pm 0.00}$   \\
        \textit{HackAttend} & 100.0              & 7.3                & 2151.3                \\
        \midrule
        \textbf{SST-2}      &                    &                    &                       \\
        Baseline            & 0.6$_{\pm 0.01}$   & 143.4$_{\pm 0.00}$ & 9.9$_{\pm 2.29}$      \\
        Imp. w/o Grad       & 0.6$_{\pm 0.00}$   & 143.6$_{\pm 0.00}$ & 8.5$_{\pm 0.33}$      \\
        Grad w/o Imp.       & 27.3$_{\pm 0.00}$  & 124.8$_{\pm 0.01}$ & 9.3$_{\pm 0.00}$      \\
        \textit{HackAttend} & 27.4               & 123.6              & 9.3                   \\
        \bottomrule
    \end{tabular}
    \caption{Experimental results for various tasks (mean $\pm$ variance over three seeds). ``Baseline'' indicates random layer, head and SA units selection, while \tx{Imp.} represent layer and head selection using Importance Score. \tx{Grad} represent SA unit selection using gradient.}
    \label{tbaseline}
\end{table}

\paragraph{Results on Different Components}
Table~\ref{tbaseline} provides a summary of baseline experiments, encompassing both the performance of the \textit{HackAttend} algorithm and the impact of each component. For our baseline, we employ random selection for layers, heads, and SA units.
The results show that incorporating importance scores indeed enhances the perturbation efficiency (lesser number of queries). However, empirically, we find that gradient alone is sufficient to identify vulnerable SA units \citep{wu2022adversarial}.
% {The "Imp. w/o GAIR" configuration moderately improves ASR by 3.22\% (p < 0.05) and reduces queries by 13.86\% (p < 0.05) compared to the baseline. In contrast, "GAIR w/o Imp." significantly outperforms the baseline with a substantial 24.83\% (p < 0.01) ASR improvement and an impressive 74.72\% (p < 0.01) reduction in queries, demonstrating its superior performance in both aspects.}

\begin{figure}[]
    \centering
    \includegraphics[width=1\linewidth]{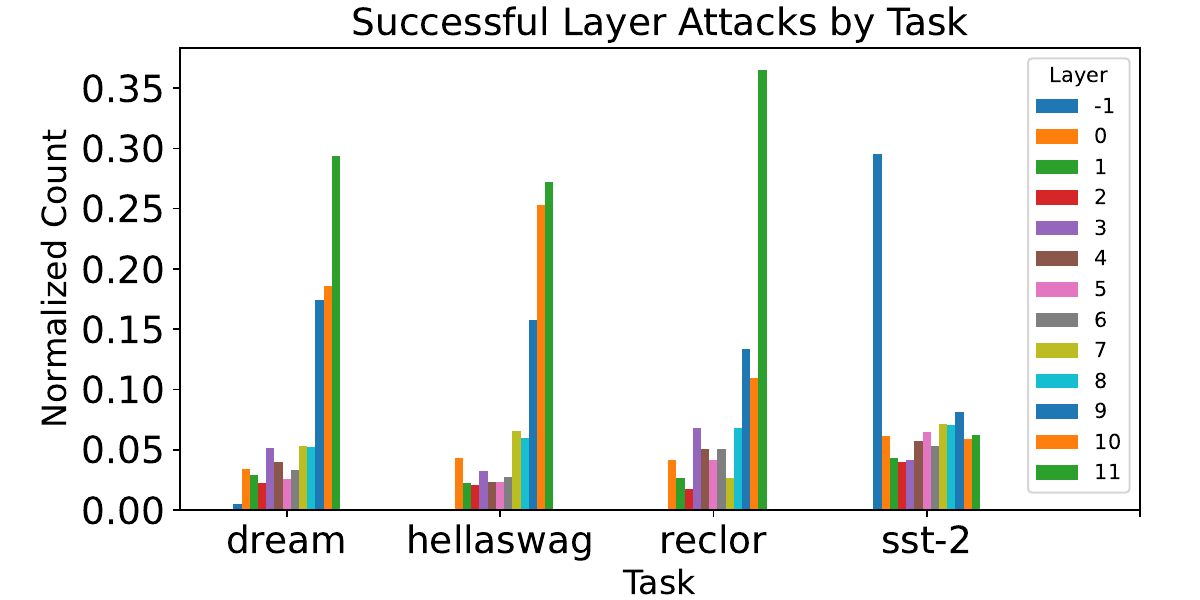}
    \caption{Normalized count of successful perturbation grouped by layers chosen and task. \textsc{-1} indicates a failed perturbation.}
    \label{fig:successful_layer_attacks_by_task}
\end{figure}

\paragraph{Vulnerability of Different Layers}
\label{sec:vulnerability_layers}
Figure~\ref{fig:successful_layer_attacks_by_task} offers valuable insight into the vulnerability of various layers. It shows the frequency of each layer was selected for perturbation. The counts are normalized to provide relative comparison of the frequencies of successful perturbations across different layers and tasks.
Higher layers, i.e. the 12\textsuperscript{th} layer, is notably more sensitive to \textit{HackAttend}, given its responsibility in managing long-range dependencies and global context.

% cumulative layer graph

% last attacked layer to success graph
% \begin{figure}[]
%     \centering
%     \includegraphics[width=1\linewidth]{figures/successful_layer_attacks_by_task.pdf}
%     \caption{Normalized count of successful attacks grouped by layers attacked and task. \textsc{-1} indicates a failed attack.}
%     \label{fig:successful_layer_attacks_by_task}
% \end{figure}

This consistency applies to all tasks except \textsc{SST-2}, where all layers exhibit similar performance. This is because sentiment analysis tasks rely heavily on local context, like keywords features, which is often sufficient to yield accurate predictions.

\subsection{Case Study}
\begin{figure}[h]
    \centering
    \begin{subfigure}[b]{0.23\textwidth}
        \centering
        \includegraphics[width=\textwidth]{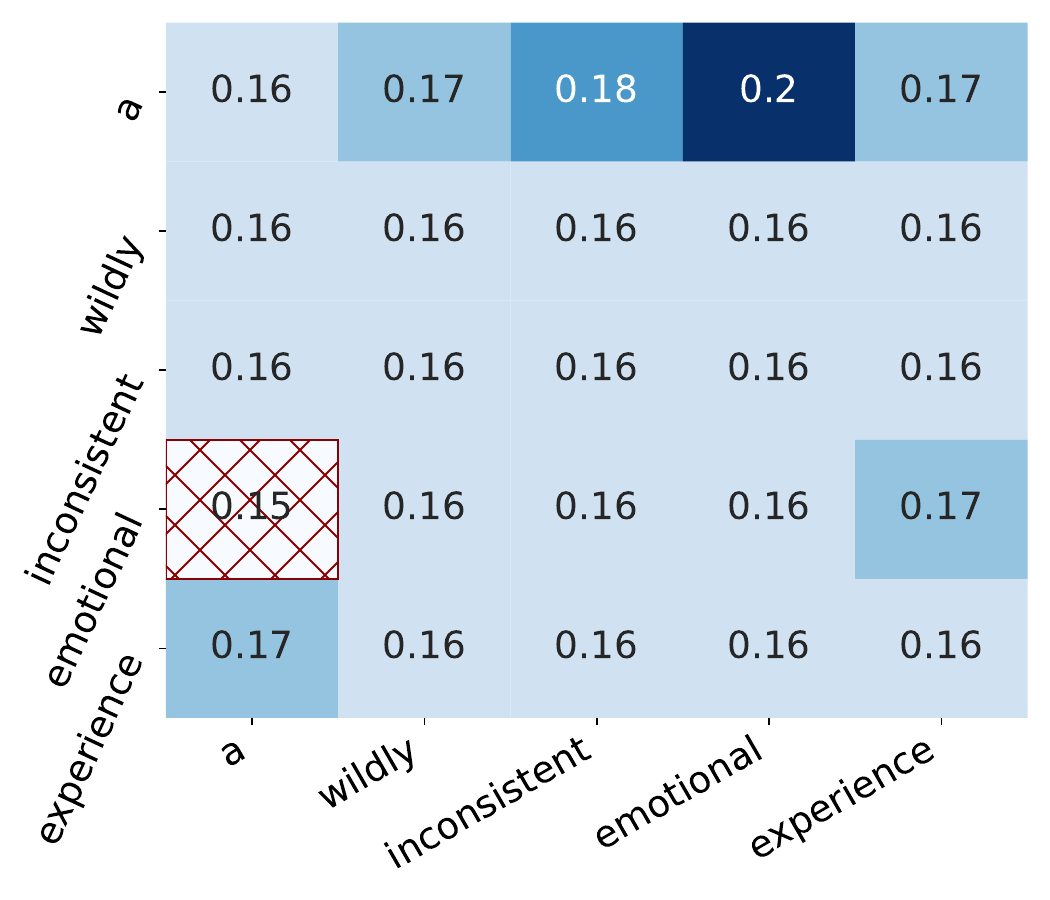}
        \caption{Original (Negative)}
        \label{fig:case_ori}
    \end{subfigure}
    \hfill
    \begin{subfigure}[b]{0.23\textwidth}
        \centering
        \includegraphics[width=\textwidth]{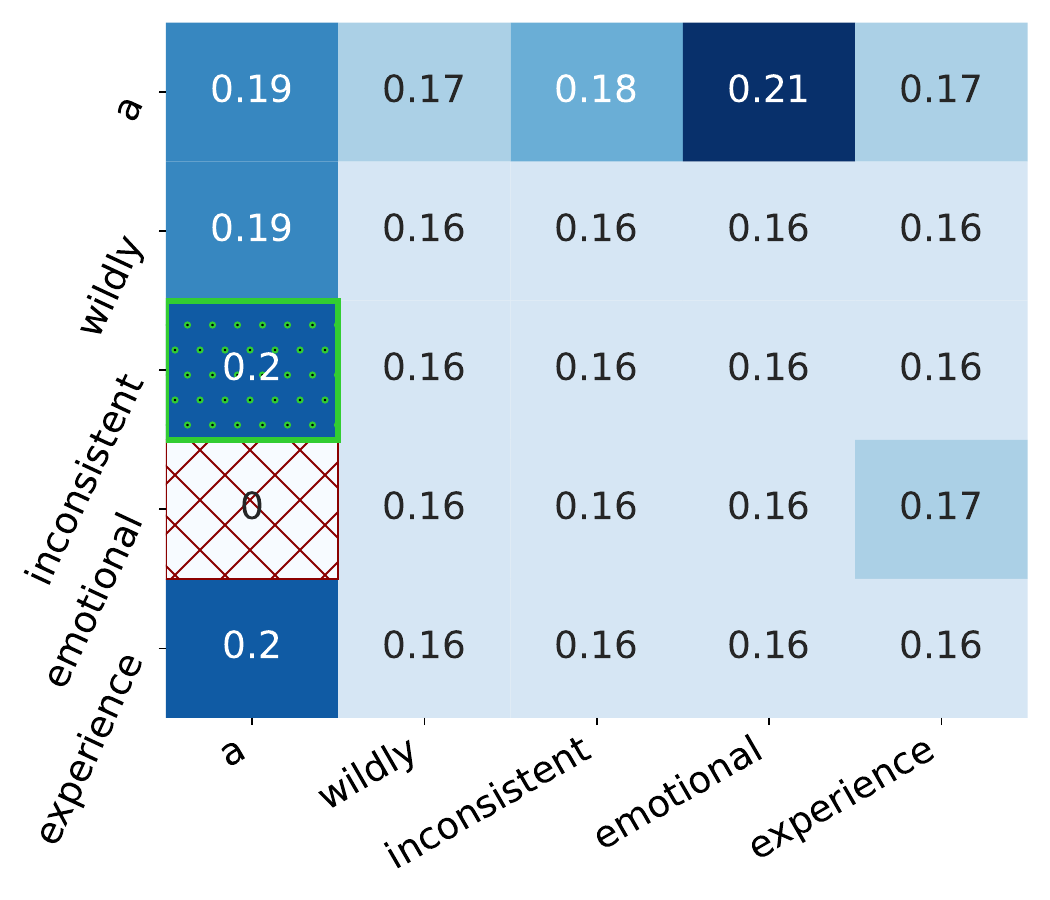}
        \caption{\textit{HackAttend} (Positive)}
        \label{fig:case_hack}
    \end{subfigure}
    \caption{Comparison of attention maps on the 2\textsuperscript{nd} layer, 10\textsuperscript{th} head of \texttt{BERT-base} before and after perturbation. The sample is "\textit{A wildly inconsistent emotional experience.}" from \textsc{SST-2}. The sentence is classified as positive after the perturbation.}
\end{figure}
The functionality of \textit{HackAttend} is illustrated through a successful example from the \textsc{SST-2} dataset, where the ground truth label is Negative (Figure~\ref{fig:case_ori}). \textit{HackAttend} masks the SA unit corresponding to the word pair \tx{a-emotional} based on gradients (labeled in \textcolor{red}{red}). This redirects the model's attention from the more relevant word pair \tx{a-emotional} to \tx{a-inconsistent} (labeled in \textcolor{green}{green}), effectively deceiving the model into focusing on a less informative feature (Figure~\ref{fig:case_hack}).

While \tx{a-inconsistent} contributes to understanding of global semantics, the word pair \tx{a-emotional} contains local semantics in the form of \textbf{emotional cues}. \textit{HackAttend} effectively misleads the model by selectively suppressing these cues through attention scores manipulation.

\section{Efficiency Analysis}

We choose two representative text attackers BERT-Attack (BA) \citep{li2020bertattack} and TextFooler (TF) \citep{DBLP:conf/aaai/JinJZS20} from TextAttack\footnote{\url{https://github.com/QData/TextAttack}} on \textsc{DREAM}, \textsc{ReClor} and \textsc{SST-2} where the benchmark results on the first 100 successful perturbations are summarized in Table~\ref{tab:efficiency_analysis}.
\begin{table}[h]
    \small
    \setlength{\tabcolsep}{3pt}
    \centering
    % \begin{tabular}{c|c|ccc}
    \begin{tabular}{@{}llcc@{}}
        \toprule
        Dataset                 & Perturbation/Attack                     & Avg Time $(s)$ & \# Query      \\ \toprule
        \multirow{3}{*}{DREAM}  & \textit{HackAttend} (ours) & \textbf{0.7}   & \textbf{13.1} \\
                                & TextFooler                 & $0.9$          & $90.7$        \\
                                & BERT-Attack                & $0.8$          & $64.3$        \\ \midrule
        \multirow{3}{*}{ReClor} & \textit{HackAttend} (ours) & \textbf{0.6}   & \textbf{7.3}  \\
                                & TextFooler                 & $1.9$          & $170.4$       \\
                                & BERT-Attack                & $2.0$          & $103.8$       \\ \midrule
        \multirow{3}{*}{SST-2}  & \textit{HackAttend} (ours) & $23.8$         & $67.2$        \\
                                & TextFooler                 & \textbf{0.4}   & $92.5$        \\
                                & BERT-Attack                & $1.0$          & \textbf{40.0} \\ \bottomrule

        % % \multirow{3}{*}{$\alpha$NLI} & 1.00      & 99.9     & 53.05      & 37.11    \\
        % %                              & 0.10      & 99.85    & 39.93      & 42.94    \\
        % %                              & 0.01      & 99.85    & 38.92      & 43.44    \\ \hline
        % % \multirow{3}{*}{MNLI (mm)}   & 1.00      & 100      & 17.37      & 37.87    \\
        % %                              & 0.10      & 99.99    & 14.59      & 41.11    \\
        % %                              & 0.01      & 99.99    & 14.46      & 41.4     \\ \hline
    \end{tabular}
    \caption{Efficiency Analysis - Average Time and Query Count for success samples. Our method outperforms TextFooler and BERT-Attack in terms of execution time and query efficiency. Candidate size (k) for BERT-Attack is set to 10.}
    \label{tab:efficiency_analysis}
\end{table}

For complex tasks, \textit{HackAttend} outperforms its counterparts in generating adversarial samples with higher efficiency. Unlike the other two algorithms, which require more time as input sequences grow longer, \textit{HackAttend} maintains relatively consistent efficiency across different sequence lengths, with the exception of SST-2.

\section{Ablation Study}
\label{sec:ablation}
All experiments use the same hyperparameter as in Section~\ref{sec:results} unless otherwise specified.

\subsection{Effect of Masking Percentage}

We evaluate the effect of mask percentage ($\alpha$) by comparing $\alpha$=\{1\%, 0.1\%, 0.01\%\}.
Table~\ref{tab:effect_of_percentage_on_asr} demonstrates that ASR maintains reasonable high, \textsc{DREAM} ($\sim 72\%$), \textsc{HellaSWAG} ($\sim 92.5\%$) and \textsc{ReClor} ($\sim84.9\%$) even with \ac{mask-percent} as low as 0.01\%. This suggests that selection of SA unit using gradients is effective in generating adversarial samples across various datasets and level of masking.
\begin{table}[h]
    \small
    \setlength{\tabcolsep}{3pt}
    \centering
    % \begin{tabular}{c|c|ccc}
    \begin{tabular}{@{}lcccc@{}}
        \toprule
        Dataset                    & Mask$\%$ & ASR$\%$ & Hamming  & \# Query \\ \toprule
        \multirow{3}{*}{DREAM}     & $1.00$   & $98.9$  & $611.4$  & $18.6$   \\
                                   & $0.10$   & $91.2$  & $62.4$   & $36.3$   \\
                                   & $0.01$   & $72.7$  & $5.7$    & $58.9$   \\ \midrule
        \multirow{3}{*}{HellaSWAG} & $1.00$   & $99.9$  & $1221.2$ & $8.8$    \\
                                   & $0.10$   & $98.9$  & $121.6$  & $17.6$   \\
                                   & $0.01$   & $92.5$  & $11.5$   & $29.7$   \\ \midrule
        \multirow{3}{*}{SST-2}     & $1.00$   & $27.4$  & $9.3$    & $123.6$  \\
                                   & $0.10$   & $6.4$   & $1.1$    & $139.0$  \\
                                   & $0.01$   & $6.4$   & $1.0$    & $139.1$  \\ \midrule

        \multirow{3}{*}{ReClor}    & $1.00$   & $100.0$ & $2151.3$ & $7.3$    \\
                                   & $0.10$   & $100.0$ & $213.3$  & $15.9$   \\
                                   & $0.01$   & $84.9$  & $19.4$   & $40.7$   \\ \bottomrule
    \end{tabular}
    \caption{Impact of masking percentage on performance metrics across different tasks.}
    \label{tab:effect_of_percentage_on_asr}
\end{table}

% Intuitively, reducing \acs{mask-percent} increases the number of necessary queries, as smaller \acs{mask-percent} values require more SA matrix perturbations.
% as depicted in Figure~\ref{fig:effect_of_perturbation_on_num_query_hamming}. 
% TODO write more on impressive results even in low mask rate
% \begin{figure}[]
%   \centering
%   \includegraphics[width=0.8\linewidth]{figures/effect_of_perturbation_on_query_hamming.pdf}
%   \caption{Effect of mask percentage on numbers of queries needed and Hamming Distance.}
%   \label{fig:effect_of_perturbation_on_num_query_hamming}
% \end{figure}

\subsection{Effect of Ranking for SA unit}
\label{sub:effect_of_importance_ranking_strategy_for_sa_unit}
Table~\ref{t3} shows both ranking by score and by negative gradient can achieve $100\%$ ASR, with some trade-off on number of queries and per-query time.
Indeed, the gradients pointing in the steepest descent direction have significant impact on the loss function, are efficient in identifying vulnerable SA units. Masking them will more likely contribute to errors in the model's predictions.

\begin{table}[h]
    \small
    \centering
    \setlength{\tabcolsep}{3pt}
    \begin{tabular}{@{}lcccc@{}}
        \toprule
        % \begin{tabular}{c|c|cccc}
        Ranking  & ASR$\%$ & Hamming  & \# Query     & \ Avg Time (s) \\ \toprule
        Gradient & $100$   & $2151.3$ & \textbf{7.3} & $0.11$         \\
        Score    & $100$   & $2151.3$ & $22.8$       & \textbf{0.08}  \\
        % grad       & $97.0$         & \textbf{581.7} & $30.6$        \\
        % magnitude  & $99.6$         & 605.0          & $22.7$        \\ 
        \bottomrule
    \end{tabular}
    \caption{Performance metrics reported on \textsc{ReClor} for different Ranking Strategies ($\alpha=1.0\%$). Score represent Attention Score. }
    \label{t3}
\end{table}

Alternatively, the attention score, assigned as a normalized value, shows the relevance of each SA unit to the model's predictions. However, since the attention score evaluates each unit individually, it may require perturbing more SA matrices to disrupt the necessary path for accurate predictions.

\section{HackAttend Inspired Smoothing}

In this section, we explore \textit{S-Attend}, a smoothing technique to improve model robustness performance, which is particularly effective in situations where attacker diversity or characteristics are not fully known.

Adversarial training (AT) \citep{goodfellow2014explaining} is proven to be an effective defense technique by augmenting the training data with adversarial samples.
However, storing such adversarial structures (e.g. for multiple attention heads/layers) \fixes{rebuttal}{slows down training or} requires large amount of storage.
Rather, we propose an efficient technique - randomly smoothing the attention weights during training - \textit{S-Attend}.
Concretely, we randomly apply masking on attention weights following a parametric Bernoulli distribution with $\alpha$ = \{0.1,0.2,0.5\} for all heads.
% Readers may refer to Table~\ref{tab:hyperparameters-sattend} for hyperparameter settings.

\fixes{rebuttal}{Table~\ref{t4} presents the performance of baseline BERT model, \textit{S-Attend}, adversarial training techniques on BERT Model (CreAT \citep{DBLP:conf/iclr/WuLSZZ23} \& FreeLB \citep{DBLP:conf/iclr/ZhuCGSGL20}) and ADA (adversarial data augmentation) technique against BERT-Attack (BA) and TextFooler (TF) adversarial attacks. ADA models refer to BERT models trained on datasets augmented by specific attack algorithms, subsequently evaluated for their defensive effectiveness against those attacks.}
% We establish a baseline using BERT-Attack and TextFooler, presenting evaluation results in Table~\ref{t4} for clean and augmented test sets in two tasks. The ADA (adversarial data augmentation) train set is generated by augmenting training data with methods from these attackers and evaluated on ADA test set. Additionally, we compare the performance of \textit{S-Attend} to FreeLB \citep{DBLP:conf/iclr/ZhuCGSGL20} and CreAT \citep{DBLP:conf/iclr/WuLSZZ23}, which are both regular adversarial training algorithms.
The column "\textit{clean\%}" and "\textit{robust\%}" represent evaluation result on clean dataset and adversarial augmented test sets using respective attackers, respectively.

\begin{figure*}[h]
    \centering
    \begin{subfigure}[b]{0.32\textwidth}
        \centering
        \includegraphics[width=\textwidth,]{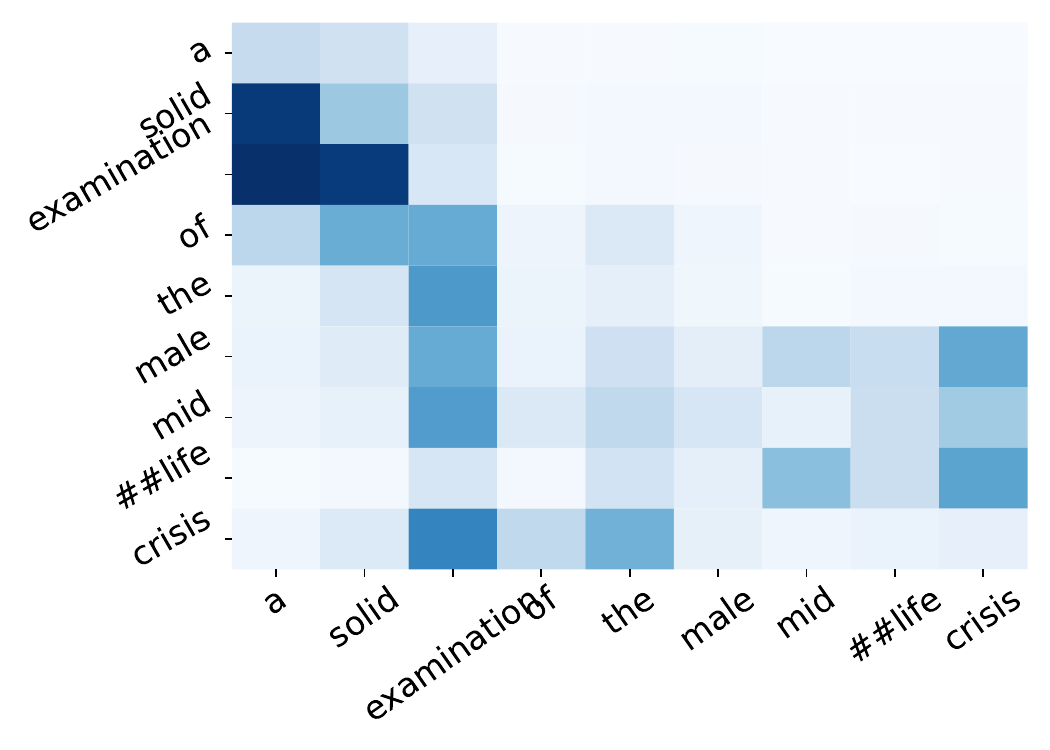}
        \caption{Normal model (clean)}
    \end{subfigure}
    \hfill
    \begin{subfigure}[b]{0.32\textwidth}
        \centering
        \includegraphics[width=\textwidth]{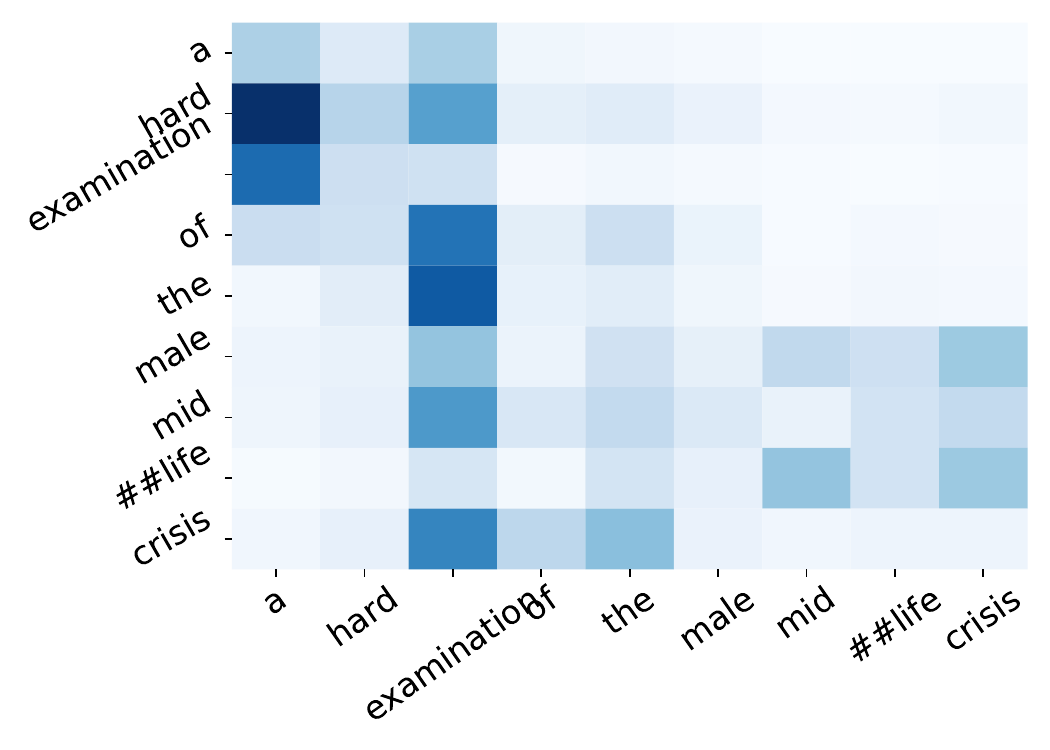}
        \caption{Normal model (augmented)}
    \end{subfigure}
    \hfill
    \begin{subfigure}[b]{0.32\textwidth}
        \centering
        \includegraphics[width=\textwidth]{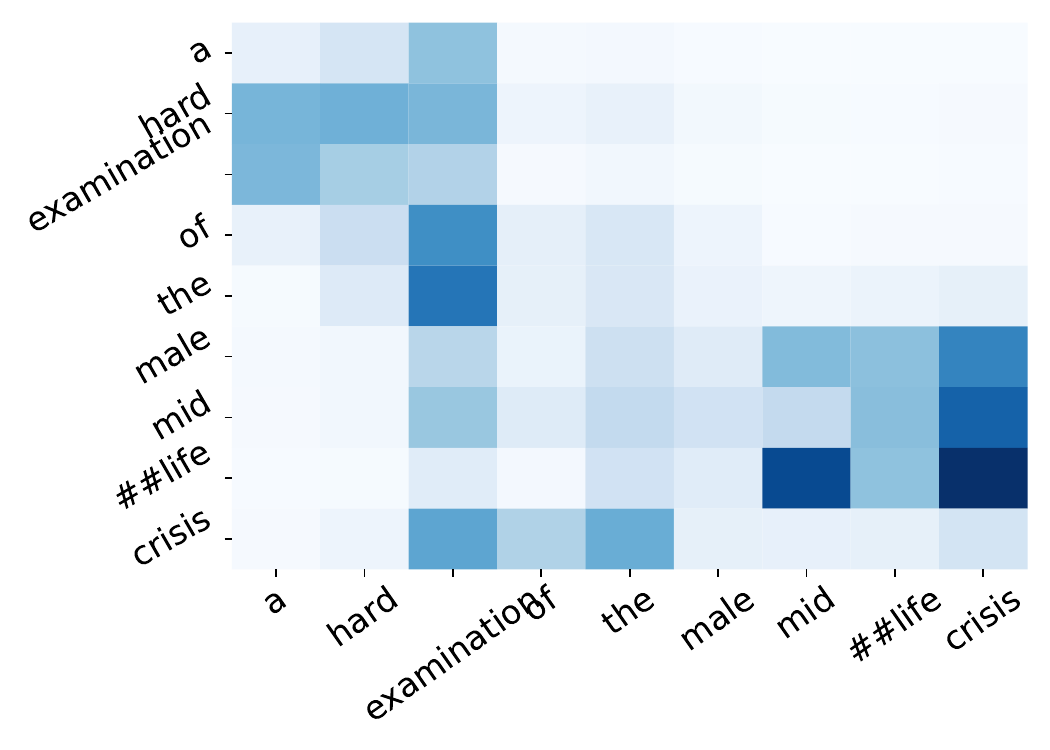}
        \caption{S-Attend model (augmented)}
    \end{subfigure}
    \caption{Comparison of attention maps for the text \textit{a solid examination of the male midlife crisis.} from \textsc{SST-2} on the 10\textsuperscript{th} layer, 4\textsuperscript{th} head of \texttt{BERT-base} between normal model (on clean and BERT-Attack augmented dataset), and \textit{S-Attend} model (on augmented dataset). After the adversarial attack, the normal model misclassifies as negative, while \textit{S-Attend} model correctly maintains as positive.}
    \label{fig:case_study_sattend}
\end{figure*}

While this smoothing technique is relatively inexpensive and straightforward, empirical experiments demonstrates that \textit{S-Attend} trained model has shown comparable or even superior robustness performance in specific tasks when compared to the regular adversarial training method, often with little to no compromise to clean performance. 

\begin{table}[h]
    \centering
    \small
    \setlength\tabcolsep{3pt}
    \begin{tabular}{@{}lrccc@{}}
        \toprule
        \multirow{2}{*}{Dataset} & \multirow{2}{*}{Defense/Smoothing}        & \multirow{2}{*}{\textit{clean}\%} & \multicolumn{2}{c}{\textit{robust}\%}                 \\
                                 &                                           &                                   & TF                                    & BA            \\ \midrule

        \multirow{8}{*}{ReClor}  & Baseline                                  & $51.8$                            & $0.8$                                 & $2.0$         \\

                                 & CreAT                                     & $49.0$                            & $46.6$                                & $48.0$        \\
                                 & FreeLB                                    & $50.4$                            & $50.2$                                & $49.6$        \\
                                 & TF(ADA)                                   & $47.8$                            & $47.4$                                & $47.8$        \\
                                 & BA(ADA)                                   & $47.4$                            & $47.0$                                & $46.6$        \\
        % training with clean samples
                                 & \textit{S-Attend} ($\alpha=0.1$)          & $48.6$                            & $47.4$                                & $47.8$        \\
                                 & \textit{S-Attend} ($\alpha=0.2$)          & $51.0$                            & $50.0$                                & $49.6$        \\
                                 & \textit{S-Attend}\textsuperscript{$\dagger$} ($\alpha=0.5$) & \textbf{52.8}                     & \textbf{51.4}                         & \textbf{51.2} \\
        % training with clean samples end
        % training with only adv
        % & \textit{S-Attend} ($\alpha=0.1$)          & $46.6$                            & $46.2$                                & $46.2$        \\
        % & \textit{S-Attend}$\dagger$ ($\alpha=0.2$) & $48.2$                            & $47.2$                                & $47.2$        \\
        % & \textit{S-Attend}($\alpha=0.5$)           & $39.2$                            & $39.0$                                & $38.8$        \\
        % training with only adv end
        \midrule
        % baseline from textattack
        \multirow{8}{*}{DREAM}   & Baseline                                  & $64.7$                            & $19.3$                                & $3.8$         \\
                                 & CreAT                                     & $65.0$                            & $55.1$                                & \textbf{55.2} \\
        % & SMART                            & 66.4                              & 58.4                                  & 56.8         \\
                                 & FreeLB                                    & \textbf{65.1}                     & \textbf{56.2}                         & $55.1$        \\
                                 & TF(ADA)                                   & $57.4$                            & $55.6$                                & $54.1$        \\
                                 & BA(ADA)                                   & $55.7$                            & $51.9$                                & $52.5$        \\
        % training with clean samples
                                 & \textit{S-Attend} ($\alpha=0.1$)          & $63.2$                            & $54.2$                                & $52.8$        \\
                                 & \textit{S-Attend}\textsuperscript{$\dagger$} ($\alpha=0.2$) & $64.4$                            & $54.6$                                & $53.7$        \\
                                 & \textit{S-Attend} ($\alpha=0.5$)          & $63.0$                            & $53.0$                                & $52.8$        \\
        % training with clean samples end
        % training with only adv
        % & \textit{S-Attend}$\dagger$ ($\alpha=0.1$) & $64.0$                            & $56.2$                                & $54.6$        \\
        % & \textit{S-Attend} ($\alpha=0.2$)          & $63.2$                            & $56.0$                                & $55.0$        \\
        % & \textit{S-Attend} ($\alpha=0.5$)          & $55.9$                            & $49.2$                                & $48.4$        \\
        % training with only adv end
        % \midrule
        % \multirow{2}{*}{HellaSwag} & Baseline                                  & $39.6$                            & $0.3$                                 & $2.1$         \\
        % %    & CreAT                                     & \textbf{25.5}                     & $27.1$                                & \textbf{28.1} \\
        %                            & FreeLB                                    & $25.7$                            & $30.4$                                & $30.6$        \\
        % training with only adv
        %  & \textit{S-Attend}$\dagger$ ($\alpha=0.1$) & $39.1$                            & $27.9$                                & $28.9$        \\
        %  & \textit{S-Attend} ($\alpha=0.2$)          & $38.4$                            & $28.4$                                & $29.5$        \\
        %  & \textit{S-Attend} ($\alpha=0.5$)          & $36.2$                            & $30.7$                                & $31.3$        \\
        % training with only adv end
        \bottomrule
    \end{tabular}
    \caption{Robust Performance Evaluation: Baseline model vs. Adversarial Training models vs. ADA models. Baseline represents regular fine-tuned BERT base. $\dagger$ denotes the best \textit{S-Attend} Model.}
    \label{t4}
\end{table}

In \textsc{DREAM}, \textit{S-Attend}$\dagger$ outperformed BA(ADA) with a 1.2\% increase in robust accuracy and a marginal 0.3\% drop in clean accuracy compared to TF(ADA) while maintaining competitive clean accuracy. In \textsc{ReClor}, the \textit{S-Attend}$\dagger$ model stands out prominently with remarkable performance. It boosts clean accuracy by 1.0\% compared to the baseline, along with reducing robust accuracy by 0.4\% and 0.6\% in both ADA test sets.

% Additionally, \textit{S-Attend} is assessed on the HANS \citep{DBLP:conf/acl/McCoyPL19} and PAWS\textsubscript{Wiki} \citep{DBLP:conf/naacl/ZhangBH19}, two datasets designed for testing model robustness in the presence of fuzzy words. Results in Table~\ref{t5} suggests that \textit{S-Attend}$\dagger$ is effective in mitigating syntactic biases within the data, achieving a 1.4\% and 1.3\% increase in accuracy, respectively.

The effectiveness of the smoothing mechanism employed by \textit{S-Attend} can be attributed to two key factors. 1)During training, \textit{S-Attend} promotes the activation of various components, which in turn aids in reducing sensitivity to noisy input data. 2) Attention weight perturbation during training helps the model to learn a more generalized representation, capturing global semantics and intricate word relationships. Fundamentally, this significantly enhances the model's robustness \citep{DBLP:conf/nips/WuX020}.
% TODO CITE

\subsection{S-Attend Against HackAttend}
To further validate \textit{S-Attend}'s effectiveness, we compared various methods against \textit{HackAttend} perturbations, as seen in Table~\ref{tab:after-hackattend}. These results highlight our model's robustness against a broad spectrum of attacks, be it conventional adversarial attacks or perturbations like \textit{HackAttend}. Notably, the smoothing technique can either reduce ASR\% or, at the very least, increase query counts.

Both CreAT and FreeLB exhibits limited effectiveness against \textit{HackAttend}, specifically on \textsc{ReClor}. In the other hand, our simple yet straightforward smoothing technique effectively mitigates this limitation. However, it's worth noting that the ASR\% of \textit{HackAttend} remains high after smoothing, indicating the SA component still presents vulnerabilities.

\begin{table}[h]
    \small
    \centering
    \begin{tabular}{@{}lrcc@{}}
        \toprule
        Dataset                          & Method            & Mask$\%$                & ASR$\%$       \\
        \toprule

        \multirow{7}{*}{\textsc{ReClor}} & Baseline          & \multirow{4}{*}{$1.00$} & $100.0$       \\
                                         & \textit{S-Attend} &                         & $100.0$       \\
                                         & CreAT             &                         & $100.0$       \\
                                         & FreeLB            &                         & $100.0$       \\
        \cmidrule{2-4}
                                         & Baseline          & \multirow{4}{*}{$0.10$} & $99.6$        \\
                                         & \textit{S-Attend} &                         & \textbf{87.1} \\
                                         & CreAT             &                         & $100.0$       \\
                                         & FreeLB            &                         & $100.0$       \\
        \midrule
        \multirow{7}{*}{\textsc{DREAM}}  & Baseline          & \multirow{4}{*}{$1.00$} & $98.9$        \\
                                         & \textit{S-Attend} &                         & \textbf{97.5} \\
                                         & CreAT             &                         & $100.0$       \\
                                         & FreeLB            &                         & $98.9$        \\
        \cmidrule{2-4}
                                         & Baseline          & \multirow{4}{*}{$0.10$} & $91.2$        \\
                                         & \textit{S-Attend} &                         & \textbf{85.1} \\
                                         & CreAT             &                         & $90.0$        \\
                                         & FreeLB            &                         & $88.3$        \\
        %         \midrule
        % \multirow{2}{*}{HellaSwag} & $1.00$   & $99.9$$\rightarrow$$98.3$        & $8.8$$\rightarrow$$21.7$         \\
        %                            & $0.10$   & $98.9$$\rightarrow$\textbf{84.2} & $17.6$$\rightarrow$\textbf{49.6} \\
        % creat                      & $1.00$   & $100.0$                          & $7.0$                            \\
        % creat                      & $0.10$   & $99.8$                           & $13.2$                           \\
        % freelb                     & $1.00$   & $99.9$                           & $10.5$                           \\
        % freelb                     & $0.10$   & $98.0$                           & $21.0$                           \\
        \bottomrule
    \end{tabular}
    \caption{Robustness evaluation against \textit{HackAttend} perturbations. Adversarial Training vs. \textit{S-Attend} smoothing. Baseline represents regular fine-tuned model. }
    \label{tab:after-hackattend}
\end{table}

% \vspace{-0.38em} with positive ground truth label 
\subsection{Case Study}
The aftereffects of applying \textit{S-Attend} could be demonstrated through a successful scenario where our model correctly maintains its predictions even when subjected to adversarial attacks. As an example from \textsc{SST-2} (Figure~\ref{fig:case_study_sattend}), the attention heatmap indicates that the words \tx{solid} and \tx{examination} are pivotal in the model's positive sentiment prediction. However, a synonym word swap (\tx{solid}$\rightarrow$\tx{hard}) by BERT-Attack causes misclassification due to spurious patterns, which suggest that models tend to heavily rely on word matching \citep{DBLP:conf/aaai/Hao0W021}.

In contrast, the \textit{S-Attend} model showcases its remarkable capability and adaptability in effectively focusing on critical cues like \tx{midlife} and \tx{crisis}. It demonstrates its ability to shift its attention, when confronted with potentially misleading word substitutions, such as synonyms and semantically related terms, while maintaining predictive accuracy in challenging linguistic scenarios.

% These particular words are considered pivotal in this context, as they are strongly associated with a neutral or positive sentiment. 
\section{Conclusion}
In this paper, we presented \textit{HackAttend}, a method for perturbing SA-based PLMs. Our experiments demonstrated its effectiveness in generating adversarial structures, particularly on complex tasks. \textit{HackAttend} achieves high success rates with minimal perturbation. Additionally, we proposed \textit{S-Attend}, a smoothing technique that enhances the model SA structure with minimal impact on training time, while exhibiting competitive performance against other adversarial training.

% we also propse a simple , particular effective on complex task like 
% our work highlights the importance of understanding the 

% \section*{Future Work}
\subsection*{Limitations}

This paper only studies on encoder-only architecture. Firstly, extending the applicability of \textit{HackAttend} to other model architectures, such as encoder-decoder architectures like T5 \citep{DBLP:journals/jmlr/RaffelSRLNMZLL20} or decoder-only architectures like GPT \citep{radford_improving_2018} are not studied. We did not evaluate \textit{HackAttend} on large language models due to the resource limitations.

\nocite{*}
\section{Bibliographical References}\label{sec:reference}
\bibliographystyle{lrec-coling2024-natbib-new}
\bibliography{lrec-coling2024-example}

\begin{thebibliography}{42}
\expandafter\ifx\csname natexlab\endcsname\relax\def\natexlab#1{#1}\fi

\bibitem[{Budhraja et~al.(2020)Budhraja, Pande, Nema, Kumar, and Khapra}]{DBLP:conf/emnlp/BudhrajaPNKK20}
Aakriti Budhraja, Madhura Pande, Preksha Nema, Pratyush Kumar, and Mitesh~M. Khapra. 2020.
\newblock \href {https://doi.org/10.18653/V1/2020.EMNLP-MAIN.260} {On the weak link between importance and prunability of attention heads}.
\newblock In \emph{Proceedings of the 2020 Conference on Empirical Methods in Natural Language Processing, {EMNLP} 2020, Online, November 16-20, 2020}, pages 3230--3235. Association for Computational Linguistics.

\bibitem[{Radford et~al.(2018)Radford, Narasimhan, Salimans, and Sutskever}]{radford_improving_2018}
Alec Radford, Karthik Narasimhan, Tim Salimans, and Ilya Sutskever. 2018.
\newblock \href {https://www.semanticscholar.org/paper/Improving-Language-Understanding-by-Generative-Radford-Narasimhan/cd18800a0fe0b668a1cc19f2ec95b5003d0a5035} {Improving language understanding by generative pre-training}.

\bibitem[{Vaswani et~al.(2017)Vaswani, Shazeer, Parmar, Uszkoreit, Jones, Gomez, Kaiser, and Polosukhin}]{DBLP:conf/nips/VaswaniSPUJGKP17}
Ashish Vaswani, Noam Shazeer, Niki Parmar, Jakob Uszkoreit, Llion Jones, Aidan~N. Gomez, Lukasz Kaiser, and Illia Polosukhin. 2017.
\newblock \href {https://proceedings.neurips.cc/paper/2017/hash/3f5ee243547dee91fbd053c1c4a845aa-Abstract.html} {Attention is all you need}.
\newblock In \emph{Advances in Neural Information Processing Systems 30: Annual Conference on Neural Information Processing Systems 2017, December 4-9, 2017, Long Beach, CA, {USA}}, pages 5998--6008.

\bibitem[{Wang et~al.(2022)Wang, Xu, Liu, Cheng, and Li}]{wang2022semattack}
Boxin Wang, Chejian Xu, Xiangyu Liu, Yu~Cheng, and Bo~Li. 2022.
\newblock \href {https://doi.org/10.18653/v1/2022.findings-naacl.14} {Semattack: Natural textual attacks via different semantic spaces}.
\newblock In \emph{Findings of the Association for Computational Linguistics: {NAACL} 2022, Seattle, WA, United States, July 10-15, 2022}, pages 176--205. Association for Computational Linguistics.

\bibitem[{Zhu et~al.(2020)Zhu, Cheng, Gan, Sun, Goldstein, and Liu}]{DBLP:conf/iclr/ZhuCGSGL20}
Chen Zhu, Yu~Cheng, Zhe Gan, Siqi Sun, Tom Goldstein, and Jingjing Liu. 2020.
\newblock \href {https://openreview.net/forum?id=BygzbyHFvB} {Freelb: Enhanced adversarial training for natural language understanding}.
\newblock In \emph{8th International Conference on Learning Representations, {ICLR} 2020, Addis Ababa, Ethiopia, April 26-30, 2020}. OpenReview.net.

\bibitem[{Raffel et~al.(2020)Raffel, Shazeer, Roberts, Lee, Narang, Matena, Zhou, Li, and Liu}]{DBLP:journals/jmlr/RaffelSRLNMZLL20}
Colin Raffel, Noam Shazeer, Adam Roberts, Katherine Lee, Sharan Narang, Michael Matena, Yanqi Zhou, Wei Li, and Peter~J. Liu. 2020.
\newblock \href {http://jmlr.org/papers/v21/20-074.html} {Exploring the limits of transfer learning with a unified text-to-text transformer}.
\newblock \emph{J. Mach. Learn. Res.}, 21:140:1--140:67.

\bibitem[{Lee et~al.(2022)Lee, Moon, Lee, and Song}]{DBLP:conf/icml/LeeMLS22}
Deokjae Lee, Seungyong Moon, Junhyeok Lee, and Hyun~Oh Song. 2022.
\newblock \href {https://proceedings.mlr.press/v162/lee22h.html} {Query-efficient and scalable black-box adversarial attacks on discrete sequential data via bayesian optimization}.
\newblock In \emph{International Conference on Machine Learning, {ICML} 2022, 17-23 July 2022, Baltimore, Maryland, {USA}}, volume 162 of \emph{Proceedings of Machine Learning Research}, pages 12478--12497. {PMLR}.

\bibitem[{Jin et~al.(2020)Jin, Jin, Zhou, and Szolovits}]{DBLP:conf/aaai/JinJZS20}
Di~Jin, Zhijing Jin, Joey~Tianyi Zhou, and Peter Szolovits. 2020.
\newblock \href {https://ojs.aaai.org/index.php/AAAI/article/view/6311} {Is {BERT} really robust? {A} strong baseline for natural language attack on text classification and entailment}.
\newblock In \emph{The Thirty-Fourth {AAAI} Conference on Artificial Intelligence, {AAAI} 2020, The Thirty-Second Innovative Applications of Artificial Intelligence Conference, {IAAI} 2020, The Tenth {AAAI} Symposium on Educational Advances in Artificial Intelligence, {EAAI} 2020, New York, NY, USA, February 7-12, 2020}, pages 8018--8025. {AAAI} Press.

\bibitem[{Wu et~al.(2020)Wu, Xia, and Wang}]{DBLP:conf/nips/WuX020}
Dongxian Wu, Shu{-}Tao Xia, and Yisen Wang. 2020.
\newblock \href {https://proceedings.neurips.cc/paper/2020/hash/1ef91c212e30e14bf125e9374262401f-Abstract.html} {Adversarial weight perturbation helps robust generalization}.
\newblock In \emph{Advances in Neural Information Processing Systems 33: Annual Conference on Neural Information Processing Systems 2020, NeurIPS 2020, December 6-12, 2020, virtual}.

\bibitem[{Shi et~al.(2021)Shi, Gao, Ren, Xu, Liang, Li, and Kwok}]{shi2021sparsebert}
Han Shi, Jiahui Gao, Xiaozhe Ren, Hang Xu, Xiaodan Liang, Zhenguo Li, and James~Tin{-}Yau Kwok. 2021.
\newblock \href {http://proceedings.mlr.press/v139/shi21a.html} {Sparsebert: Rethinking the importance analysis in self-attention}.
\newblock In \emph{Proceedings of the 38th International Conference on Machine Learning, {ICML} 2021, 18-24 July 2021, Virtual Event}, volume 139 of \emph{Proceedings of Machine Learning Research}, pages 9547--9557. {PMLR}.

\bibitem[{Wu and Zhao(2022)}]{wu2022adversarial}
Hongqiu Wu and Hai Zhao. 2022.
\newblock \href {https://doi.org/10.48550/arXiv.2206.12608} {Adversarial self-attention for language understanding}.
\newblock \emph{CoRR}, abs/2206.12608.

\bibitem[{Wu et~al.(2023{\natexlab{a}})Wu, Liu, Zhao, and Zhang}]{wu2023empower}
Hongqiu Wu, Linfeng Liu, Hai Zhao, and Min Zhang. 2023{\natexlab{a}}.
\newblock Empower nested boolean logic via self-supervised curriculum learning.
\newblock \emph{arXiv preprint arXiv:2310.05450}.

\bibitem[{Wu et~al.(2023{\natexlab{b}})Wu, Liu, Shi, Zhao, and Zhang}]{DBLP:conf/iclr/WuLSZZ23}
Hongqiu Wu, Yongxiang Liu, Hanwen Shi, Hai Zhao, and Min Zhang. 2023{\natexlab{b}}.
\newblock \href {https://openreview.net/pdf?id=xZD10GhCvM} {Toward adversarial training on contextualized language representation}.
\newblock In \emph{The Eleventh International Conference on Learning Representations, {ICLR} 2023, Kigali, Rwanda, May 1-5, 2023}. OpenReview.net.

\bibitem[{Wu et~al.(2023{\natexlab{c}})Wu, Liu, Shi, hai zhao, and Zhang}]{wu2023toward}
Hongqiu Wu, Yongxiang Liu, Hanwen Shi, hai zhao, and Min Zhang. 2023{\natexlab{c}}.
\newblock \href {https://openreview.net/forum?id=xZD10GhCvM} {Toward adversarial training on contextualized language representation}.
\newblock In \emph{The Eleventh International Conference on Learning Representations}.

\bibitem[{Goodfellow et~al.(2015)Goodfellow, Shlens, and Szegedy}]{goodfellow2014explaining}
Ian~J. Goodfellow, Jonathon Shlens, and Christian Szegedy. 2015.
\newblock \href {http://arxiv.org/abs/1412.6572} {Explaining and harnessing adversarial examples}.
\newblock In \emph{3rd International Conference on Learning Representations, {ICLR} 2015, San Diego, CA, USA, May 7-9, 2015, Conference Track Proceedings}.

\bibitem[{Devlin et~al.(2019)Devlin, Chang, Lee, and Toutanova}]{DBLP:conf/naacl/DevlinCLT19}
Jacob Devlin, Ming{-}Wei Chang, Kenton Lee, and Kristina Toutanova. 2019.
\newblock \href {https://doi.org/10.18653/v1/n19-1423} {{BERT:} pre-training of deep bidirectional transformers for language understanding}.
\newblock In \emph{Proceedings of the 2019 Conference of the North American Chapter of the Association for Computational Linguistics: Human Language Technologies, {NAACL-HLT} 2019, Minneapolis, MN, USA, June 2-7, 2019, Volume 1 (Long and Short Papers)}, pages 4171--4186. Association for Computational Linguistics.

\bibitem[{Ebrahimi et~al.(2018)Ebrahimi, Rao, Lowd, and Dou}]{ebrahimi2018hotflip}
Javid Ebrahimi, Anyi Rao, Daniel Lowd, and Dejing Dou. 2018.
\newblock \href {https://doi.org/10.18653/v1/P18-2006} {Hotflip: White-box adversarial examples for text classification}.
\newblock In \emph{Proceedings of the 56th Annual Meeting of the Association for Computational Linguistics, {ACL} 2018, Melbourne, Australia, July 15-20, 2018, Volume 2: Short Papers}, pages 31--36. Association for Computational Linguistics.

\bibitem[{Gao et~al.(2018)Gao, Lanchantin, Soffa, and Qi}]{gao2018black}
Ji~Gao, Jack Lanchantin, Mary~Lou Soffa, and Yanjun Qi. 2018.
\newblock \href {https://doi.org/10.1109/SPW.2018.00016} {Black-box generation of adversarial text sequences to evade deep learning classifiers}.
\newblock In \emph{2018 {IEEE} Security and Privacy Workshops, {SP} Workshops 2018, San Francisco, CA, USA, May 24, 2018}, pages 50--56. {IEEE} Computer Society.

\bibitem[{Li et~al.(2019)Li, Ji, Du, Li, and Wang}]{li2018textbugger}
Jinfeng Li, Shouling Ji, Tianyu Du, Bo~Li, and Ting Wang. 2019.
\newblock \href {https://www.ndss-symposium.org/ndss-paper/textbugger-generating-adversarial-text-against-real-world-applications/} {Textbugger: Generating adversarial text against real-world applications}.
\newblock In \emph{26th Annual Network and Distributed System Security Symposium, {NDSS} 2019, San Diego, California, USA, February 24-27, 2019}. The Internet Society.

\bibitem[{Sun et~al.(2019)Sun, Yu, Chen, Yu, Choi, and Cardie}]{DBLP:journals/tacl/SunYCYCC19}
Kai Sun, Dian Yu, Jianshu Chen, Dong Yu, Yejin Choi, and Claire Cardie. 2019.
\newblock \href {https://doi.org/10.1162/tacl\_a\_00264} {{DREAM:} {A} challenge dataset and models for dialogue-based reading comprehension}.
\newblock \emph{Trans. Assoc. Comput. Linguistics}, 7:217--231.

\bibitem[{Li et~al.(2020)Li, Ma, Guo, Xue, and Qiu}]{li2020bertattack}
Linyang Li, Ruotian Ma, Qipeng Guo, Xiangyang Xue, and Xipeng Qiu. 2020.
\newblock \href {https://doi.org/10.18653/v1/2020.emnlp-main.500} {{BERT-ATTACK:} adversarial attack against {BERT} using {BERT}}.
\newblock In \emph{Proceedings of the 2020 Conference on Empirical Methods in Natural Language Processing, {EMNLP} 2020, Online, November 16-20, 2020}, pages 6193--6202. Association for Computational Linguistics.

\bibitem[{Zhou et~al.(2020)Zhou, Duan, Liu, and Shum}]{ZHOU2020275}
Ming Zhou, Nan Duan, Shujie Liu, and Heung-Yeung Shum. 2020.
\newblock \href {https://doi.org/https://doi.org/10.1016/j.eng.2019.12.014} {Progress in neural nlp: Modeling, learning, and reasoning}.
\newblock \emph{Engineering}, 6(3):275--290.

\bibitem[{Alzantot et~al.(2018)Alzantot, Sharma, Elgohary, Ho, Srivastava, and Chang}]{alzantot2018generating}
Moustafa Alzantot, Yash Sharma, Ahmed Elgohary, Bo{-}Jhang Ho, Mani~B. Srivastava, and Kai{-}Wei Chang. 2018.
\newblock \href {https://doi.org/10.18653/v1/d18-1316} {Generating natural language adversarial examples}.
\newblock In \emph{Proceedings of the 2018 Conference on Empirical Methods in Natural Language Processing, Brussels, Belgium, October 31 - November 4, 2018}, pages 2890--2896. Association for Computational Linguistics.

\bibitem[{Srivastava et~al.(2014)Srivastava, Hinton, Krizhevsky, Sutskever, and Salakhutdinov}]{DBLP:journals/jmlr/SrivastavaHKSS14}
Nitish Srivastava, Geoffrey~E. Hinton, Alex Krizhevsky, Ilya Sutskever, and Ruslan Salakhutdinov. 2014.
\newblock \href {https://doi.org/10.5555/2627435.2670313} {Dropout: a simple way to prevent neural networks from overfitting}.
\newblock \emph{J. Mach. Learn. Res.}, 15(1):1929--1958.

\bibitem[{Michel et~al.(2019)Michel, Levy, and Neubig}]{michel2019sixteen}
Paul Michel, Omer Levy, and Graham Neubig. 2019.
\newblock \href {https://proceedings.neurips.cc/paper/2019/hash/2c601ad9d2ff9bc8b282670cdd54f69f-Abstract.html} {Are sixteen heads really better than one?}
\newblock In \emph{Advances in Neural Information Processing Systems 32: Annual Conference on Neural Information Processing Systems 2019, NeurIPS 2019, December 8-14, 2019, Vancouver, BC, Canada}, pages 14014--14024.

\bibitem[{He et~al.(2021)He, Liu, Gao, and Chen}]{DBLP:conf/iclr/HeLGC21}
Pengcheng He, Xiaodong Liu, Jianfeng Gao, and Weizhu Chen. 2021.
\newblock \href {https://openreview.net/forum?id=XPZIaotutsD} {Deberta: decoding-enhanced bert with disentangled attention}.
\newblock In \emph{9th International Conference on Learning Representations, {ICLR} 2021, Virtual Event, Austria, May 3-7, 2021}. OpenReview.net.

\bibitem[{Socher et~al.(2013)Socher, Perelygin, Wu, Chuang, Manning, Ng, and Potts}]{DBLP:conf/emnlp/SocherPWCMNP13}
Richard Socher, Alex Perelygin, Jean Wu, Jason Chuang, Christopher~D. Manning, Andrew~Y. Ng, and Christopher Potts. 2013.
\newblock \href {https://aclanthology.org/D13-1170/} {Recursive deep models for semantic compositionality over a sentiment treebank}.
\newblock In \emph{Proceedings of the 2013 Conference on Empirical Methods in Natural Language Processing, {EMNLP} 2013, 18-21 October 2013, Grand Hyatt Seattle, Seattle, Washington, USA, {A} meeting of SIGDAT, a Special Interest Group of the {ACL}}, pages 1631--1642. {ACL}.

\bibitem[{Zellers et~al.(2019)Zellers, Holtzman, Bisk, Farhadi, and Choi}]{DBLP:conf/acl/ZellersHBFC19}
Rowan Zellers, Ari Holtzman, Yonatan Bisk, Ali Farhadi, and Yejin Choi. 2019.
\newblock \href {https://doi.org/10.18653/v1/p19-1472} {Hellaswag: Can a machine really finish your sentence?}
\newblock In \emph{Proceedings of the 57th Conference of the Association for Computational Linguistics, {ACL} 2019, Florence, Italy, July 28- August 2, 2019, Volume 1: Long Papers}, pages 4791--4800. Association for Computational Linguistics.

\bibitem[{Ren et~al.(2019)Ren, Deng, He, and Che}]{ren2019generating}
Shuhuai Ren, Yihe Deng, Kun He, and Wanxiang Che. 2019.
\newblock \href {https://doi.org/10.18653/v1/p19-1103} {Generating natural language adversarial examples through probability weighted word saliency}.
\newblock In \emph{Proceedings of the 57th Conference of the Association for Computational Linguistics, {ACL} 2019, Florence, Italy, July 28- August 2, 2019, Volume 1: Long Papers}, pages 1085--1097. Association for Computational Linguistics.

\bibitem[{Cao and Wang(2021)}]{DBLP:conf/naacl/CaoW21}
Shuyang Cao and Lu~Wang. 2021.
\newblock \href {https://doi.org/10.18653/V1/2021.NAACL-MAIN.397} {Attention head masking for inference time content selection in abstractive summarization}.
\newblock In \emph{Proceedings of the 2021 Conference of the North American Chapter of the Association for Computational Linguistics: Human Language Technologies, {NAACL-HLT} 2021, Online, June 6-11, 2021}, pages 5008--5016. Association for Computational Linguistics.

\bibitem[{Zhang et~al.(2020{\natexlab{a}})Zhang, Kishore, Wu, Weinberger, and Artzi}]{zhang2019bertscore}
Tianyi Zhang, Varsha Kishore, Felix Wu, Kilian~Q. Weinberger, and Yoav Artzi. 2020{\natexlab{a}}.
\newblock \href {https://openreview.net/forum?id=SkeHuCVFDr} {Bertscore: Evaluating text generation with {BERT}}.
\newblock In \emph{8th International Conference on Learning Representations, {ICLR} 2020, Addis Ababa, Ethiopia, April 26-30, 2020}. OpenReview.net.

\bibitem[{McCoy et~al.(2019)McCoy, Pavlick, and Linzen}]{DBLP:conf/acl/McCoyPL19}
Tom McCoy, Ellie Pavlick, and Tal Linzen. 2019.
\newblock \href {https://doi.org/10.18653/v1/p19-1334} {Right for the wrong reasons: Diagnosing syntactic heuristics in natural language inference}.
\newblock In \emph{Proceedings of the 57th Conference of the Association for Computational Linguistics, {ACL} 2019, Florence, Italy, July 28- August 2, 2019, Volume 1: Long Papers}, pages 3428--3448. Association for Computational Linguistics.

\bibitem[{Yu et~al.(2020)Yu, Jiang, Dong, and Feng}]{yu2020reclor}
Weihao Yu, Zihang Jiang, Yanfei Dong, and Jiashi Feng. 2020.
\newblock \href {https://openreview.net/forum?id=HJgJtT4tvB} {Reclor: {A} reading comprehension dataset requiring logical reasoning}.
\newblock In \emph{8th International Conference on Learning Representations, {ICLR} 2020, Addis Ababa, Ethiopia, April 26-30, 2020}. OpenReview.net.

\bibitem[{You et~al.(2020)You, Sun, and Iyyer}]{you2020hard}
Weiqiu You, Simeng Sun, and Mohit Iyyer. 2020.
\newblock \href {https://doi.org/10.18653/v1/2020.acl-main.687} {Hard-coded gaussian attention for neural machine translation}.
\newblock In \emph{Proceedings of the 58th Annual Meeting of the Association for Computational Linguistics, {ACL} 2020, Online, July 5-10, 2020}, pages 7689--7700. Association for Computational Linguistics.

\bibitem[{Zhao et~al.(2019)Zhao, Peyrard, Liu, Gao, Meyer, and Eger}]{zhao2019moverscore}
Wei Zhao, Maxime Peyrard, Fei Liu, Yang Gao, Christian~M. Meyer, and Steffen Eger. 2019.
\newblock \href {https://doi.org/10.18653/v1/D19-1053} {Moverscore: Text generation evaluating with contextualized embeddings and earth mover distance}.
\newblock In \emph{Proceedings of the 2019 Conference on Empirical Methods in Natural Language Processing and the 9th International Joint Conference on Natural Language Processing, {EMNLP-IJCNLP} 2019, Hong Kong, China, November 3-7, 2019}, pages 563--578. Association for Computational Linguistics.

\bibitem[{Hao et~al.(2021)Hao, Dong, Wei, and Xu}]{DBLP:conf/aaai/Hao0W021}
Yaru Hao, Li~Dong, Furu Wei, and Ke~Xu. 2021.
\newblock \href {https://doi.org/10.1609/aaai.v35i14.17533} {Self-attention attribution: Interpreting information interactions inside transformer}.
\newblock In \emph{Thirty-Fifth {AAAI} Conference on Artificial Intelligence, {AAAI} 2021, Thirty-Third Conference on Innovative Applications of Artificial Intelligence, {IAAI} 2021, The Eleventh Symposium on Educational Advances in Artificial Intelligence, {EAAI} 2021, Virtual Event, February 2-9, 2021}, pages 12963--12971. {AAAI} Press.

\bibitem[{Liu et~al.(2019)Liu, Ott, Goyal, Du, Joshi, Chen, Levy, Lewis, Zettlemoyer, and Stoyanov}]{DBLP:journals/corr/abs-1907-11692}
Yinhan Liu, Myle Ott, Naman Goyal, Jingfei Du, Mandar Joshi, Danqi Chen, Omer Levy, Mike Lewis, Luke Zettlemoyer, and Veselin Stoyanov. 2019.
\newblock \href {http://arxiv.org/abs/1907.11692} {Roberta: {A} robustly optimized {BERT} pretraining approach}.
\newblock \emph{CoRR}, abs/1907.11692.

\bibitem[{Zhang et~al.(2019)Zhang, Baldridge, and He}]{DBLP:conf/naacl/ZhangBH19}
Yuan Zhang, Jason Baldridge, and Luheng He. 2019.
\newblock \href {https://doi.org/10.18653/v1/n19-1131} {{PAWS:} paraphrase adversaries from word scrambling}.
\newblock In \emph{Proceedings of the 2019 Conference of the North American Chapter of the Association for Computational Linguistics: Human Language Technologies, {NAACL-HLT} 2019, Minneapolis, MN, USA, June 2-7, 2019, Volume 1 (Long and Short Papers)}, pages 1298--1308. Association for Computational Linguistics.

\bibitem[{Yu and Herman(2005)}]{Rubner2000earthmover}
Zhenghua Yu and Gunawan Herman. 2005.
\newblock \href {https://doi.org/10.1109/ICME.2005.1521516} {On the earth mover's distance as a histogram similarity metric for image retrieval}.
\newblock In \emph{Proceedings of the 2005 {IEEE} International Conference on Multimedia and Expo, {ICME} 2005, July 6-9, 2005, Amsterdam, The Netherlands}, pages 686--689. {IEEE} Computer Society.

\bibitem[{Lan et~al.(2020)Lan, Chen, Goodman, Gimpel, Sharma, and Soricut}]{DBLP:conf/iclr/LanCGGSS20}
Zhenzhong Lan, Mingda Chen, Sebastian Goodman, Kevin Gimpel, Piyush Sharma, and Radu Soricut. 2020.
\newblock \href {https://openreview.net/forum?id=H1eA7AEtvS} {{ALBERT:} {A} lite {BERT} for self-supervised learning of language representations}.
\newblock In \emph{8th International Conference on Learning Representations, {ICLR} 2020, Addis Ababa, Ethiopia, April 26-30, 2020}. OpenReview.net.

\bibitem[{Fan et~al.(2021)Fan, Gong, Liu, Wei, Wang, Jiao, Duan, Zhang, and Huang}]{DBLP:conf/naacl/FanGLWWJDZH21}
Zhihao Fan, Yeyun Gong, Dayiheng Liu, Zhongyu Wei, Siyuan Wang, Jian Jiao, Nan Duan, Ruofei Zhang, and Xuanjing Huang. 2021.
\newblock \href {https://doi.org/10.18653/V1/2021.NAACL-MAIN.135} {Mask attention networks: Rethinking and strengthen transformer}.
\newblock In \emph{Proceedings of the 2021 Conference of the North American Chapter of the Association for Computational Linguistics: Human Language Technologies, {NAACL-HLT} 2021, Online, June 6-11, 2021}, pages 1692--1701. Association for Computational Linguistics.

\bibitem[{Zhang et~al.(2020{\natexlab{b}})Zhang, Wu, Zhou, Duan, Zhao, and Wang}]{zhang2020sg}
Zhuosheng Zhang, Yuwei Wu, Junru Zhou, Sufeng Duan, Hai Zhao, and Rui Wang. 2020{\natexlab{b}}.
\newblock \href {https://ojs.aaai.org/index.php/AAAI/article/view/6511} {Sg-net: Syntax-guided machine reading comprehension}.
\newblock In \emph{The Thirty-Fourth {AAAI} Conference on Artificial Intelligence, {AAAI} 2020, The Thirty-Second Innovative Applications of Artificial Intelligence Conference, {IAAI} 2020, The Tenth {AAAI} Symposium on Educational Advances in Artificial Intelligence, {EAAI} 2020, New York, NY, USA, February 7-12, 2020}, pages 9636--9643. {AAAI} Press.

\end{thebibliography}

% TODO FOR FINAL PAPER
\section{Appendices}

\label{sec:appendix}
\subsection*{Fine-tuning Details}

\begin{table}[H]
    \centering
    \small
    \begin{tabular}{@{}lccccc@{}}
        \toprule
                  & LR   & BSZ  & EP  & WP     & MSL   \\ \midrule
        DREAM     & $3e-5$ & $16$ & $8$ & $0.1 $ & $128$ \\
        HellaSWAG & $2e-5$ & $32$ & $3$ & $0.1 $ & $128$ \\
        ReClor    & $2e-5$ & $24$ & $6$ & $0.06$ & $128$ \\
        SST-2     & $2e-5$ & $32$ & $3$ & $0.06$ & $128$ \\
        % MNLI      & 3e-5 & 32   & 3   & 0.06   & 128   \\
        % QNLI      & 2e-5 & 32  & 3  & 0.06 & 128 \\
        % QQP       & 5e-5 & 32  & 3  & 0.06 & 128 \\
        % STS-B     & 5e-5 & 16  & 3  & 0.06 & 128 \\
        % WNUT-17   & 5e-5 & 16  & 5  & 0.1  & 64  \\
        % ANLI      & 3e-5 & 32  & 3  & 0.06 & 128 \\
        % PAWS-QQP  & 5e-5 & 16  & 3  & 0.06 & 128 \\
        \bottomrule
    \end{tabular}
    \caption{Suggested fine-tuning setting. LR: learning rate; BSZ: batch size; EP: training epochs; WP: warmup proportion; MSL: sequence length;}
    \label{tab:hyperparameters}
\end{table}
% \begin{table}[H]
%     \centering
%     \small
%     \begin{tabular}{@{}lcccccc@{}}
%         \toprule
%                   & LR   & BSZ  & EP  & WP     & MSL  & $\alpha$ \\ \midrule
%         DREAM     & 2e-5 & $32$ & $8$ & $0.06 $ & $128$ & 0.2 \\
%         % DREAM (ADA on TF) & 2e-5 & 32 & 8 & 0.06 & 128 \\
%         % DREAM (ADA on BA) & 3e-5 & 32 & 8 & 0.06 & 128 \\
%         % HellaSWAG & 2e-5 & $32$ & $3$ & $0.1 $ & $128$ \\
%         ReClor    & 2e-5 & $32$ & $6$ & $0.06$ & $128$ & 0.5\\
%         % ReClor (ADA on BA) & 2e-5 & $32$ & $6$ & $0.1$ & $128$ \\
%         % ReClor (ADA on TF) & 2e-5 & $32$ & $6$ & $0.1$ & $128$ \\
%         % SST-2     & 2e-5 & $32$ & $3$ & $0.06$ & $128$ & 0.2 \\
%         % MNLI      & 3e-5 & 32   & 3   & 0.06   & 128   \\
%         % QNLI      & 2e-5 & 32  & 3  & 0.06 & 128 \\
%         % QQP       & 5e-5 & 32  & 3  & 0.06 & 128 \\
%         % STS-B     & 5e-5 & 16  & 3  & 0.06 & 128 \\
%         % WNUT-17   & 5e-5 & 16  & 5  & 0.1  & 64  \\
%         % ANLI      & 3e-5 & 32  & 3  & 0.06 & 128 \\
%         % PAWS-QQP  & 5e-5 & 16  & 3  & 0.06 & 128 \\
%         \bottomrule
%     \end{tabular}
%     \caption{Suggested \textit{S-Attend setting}. LR: learning rate; BSZ: batch size; EP: training epochs; WP: warmup proportion; MSL: sequence length; $\alpha$: mask rate}
%     \label{tab:hyperparameters-sattend}
% \end{table}

% \section{Language Resource References}
% \label{lr:ref}
% \bibliographystylelanguageresource{lrec-coling2024-natbib}
% \bibliographystylelanguageresource{anthology}
% % \bibliographylanguageresource{languageresource}

\end{document}